\documentclass[lettersize,journal]{IEEEtran}
\usepackage{amsmath,amsfonts}
\usepackage{algorithmic}
\usepackage{algorithm}
\usepackage{array}
\usepackage[caption=false,font=normalsize,labelfont=sf,textfont=sf]{subfig}
\usepackage{textcomp}
\usepackage{stfloats}
\usepackage{url}
\usepackage{verbatim}
\usepackage{graphicx}
\usepackage{cite}
\usepackage[bookmarks,colorlinks,linkcolor=red, anchorcolor=blue, citecolor=green]{hyperref}
            
\usepackage{amsmath}
\usepackage{amssymb}

\usepackage{microtype}
\usepackage{color}
\usepackage{colortbl}
\usepackage{multirow}
\usepackage{makecell}
\usepackage{hhline}
\usepackage[american]{babel}
\usepackage{bbding}
\usepackage{cuted}
\usepackage{ragged2e}
\usepackage{comment}
\usepackage{CJKutf8}

\definecolor{lightgray}{gray}{.92}
\definecolor{tinygray}{gray}{.96}

\newcommand{\etal}{\textit{et al}.}
\newcommand{\ie}{\textit{i}.\textit{e}.}
\newcommand{\eg}{\textit{e}.\textit{g}.}

\begin{document}

\title{Enhanced Generative Structure Prior for \\Chinese Text Image Super-resolution}

\author{Xiaoming Li,~\IEEEmembership{Member,~IEEE},
        Wangmeng Zuo,~\IEEEmembership{Senior Member,~IEEE},
	Chen Change Loy,~\IEEEmembership{Senior Member,~IEEE}
\thanks{
\IEEEcompsocthanksitem 
This work was supported by the RIE2020 Industry Alignment Fund Industry Collaboration Projects (IAF-ICP) Funding Initiative, as well as cash and inkind contribution from the industry partner(s).
\IEEEcompsocthanksitem Xiaoming  Li is with S-Lab, Nanyang Technological University,	Singapore. E-mail: xiaoming.li@ntu.edu.sg
		\IEEEcompsocthanksitem Wangmeng Zuo is with the Faculty of Computing, Harbin Institute of Technology, Harbin, China. E-mail: cswmzuo@gmail.com
		\IEEEcompsocthanksitem Chen Change Loy is with S-Lab, Nanyang Technological University, Singapore. E-mail: ccloy@ntu.edu.sg (Corresponding author)}
}



\maketitle

\begin{abstract}
Faithful text image super-resolution (SR) is challenging because each character has a unique structure and usually exhibits diverse font styles and layouts.
While existing methods primarily focus on English text, less attention has been paid to more complex scripts like Chinese.
In this paper, we introduce a high-quality text image SR framework designed to restore the precise strokes of low-resolution (LR) Chinese characters. Unlike methods that rely on character recognition priors to regularize the SR task, we propose a novel structure prior that offers structure-level guidance to enhance visual quality.
Our framework incorporates this structure prior within a StyleGAN model, leveraging its generative capabilities for restoration. To maintain the integrity of character structures while accommodating various font styles and layouts, we implement a codebook-based mechanism that restricts the generative space of StyleGAN. Each code in the codebook represents the structure of a specific character, while the vector $w$ in StyleGAN controls the character's style, including typeface, orientation, and location.
Through the collaborative interaction between the codebook and style, we generate a high-resolution structure prior that aligns with LR characters both spatially and structurally. Experiments demonstrate that this structure prior provides robust, character-specific guidance, enabling the accurate restoration of clear strokes in degraded characters, even for real-world LR Chinese text with irregular layouts.
Our code and pre-trained models will be available at \url{https://github.com/csxmli2016/MARCONetPlusPlus}.
\end{abstract}


\section{Introduction}\label{sec:introduction}
\IEEEPARstart{R}{eal-world} text image super-resolution (SR) aims at recovering a high-resolution (HR) image from a low-resolution (LR) text image that has undergone unknown degradation processes. This task holds significant practical value across various real-world applications, such as restoring text in scene images (\eg, road signs), historical documents (\eg, newspapers), and text captions in digital media.
Applying general image SR methods to text images often leads to distorted strokes, which can negatively affect visual clarity and potentially alter the intended meaning of the text. 
As a result, the faithful restoration of text images has become increasingly important.
However, most existing text restoration methods focus on English letters and numbers, with little attention given to more complex characters, such as Chinese.

This seemingly straightforward task actually presents numerous challenges. 
First, each Chinese character comprises complex and unique structures, such as `\begin{CJK*}{UTF8}{gbsn}整\end{CJK*}', which contains over 15 strokes. Existing text SR methods may easily disrupt these structures by introducing distorted or extra strokes, especially when encountering severe degradation. Such distortions can easily alter the meaning, as many characters appear similar yet have entirely different meanings, like `\begin{CJK*}{UTF8}{bsmi}已\end{CJK*}' and `\begin{CJK*}{UTF8}{bsmi}己\end{CJK*}', which mean `already' and `self', respectively.
Second, although each character has a fixed structure, it often appears differently in various font families. When introducing the external character structure for a guided restoration, it is essential to ensure consistency in the typeface between the LR input and the HR-guided image.
Third, real-world text images have irregular layouts, appearing in perspective or curved arrangements. 
Faithful restoration of these characters requires precise alignment of orientation and location.

\begin{figure*}[!t]
    \setlength{\abovecaptionskip}{4pt}
    \setlength{\belowcaptionskip}{-12pt}
    \centering
    \includegraphics[width=.84\textwidth]{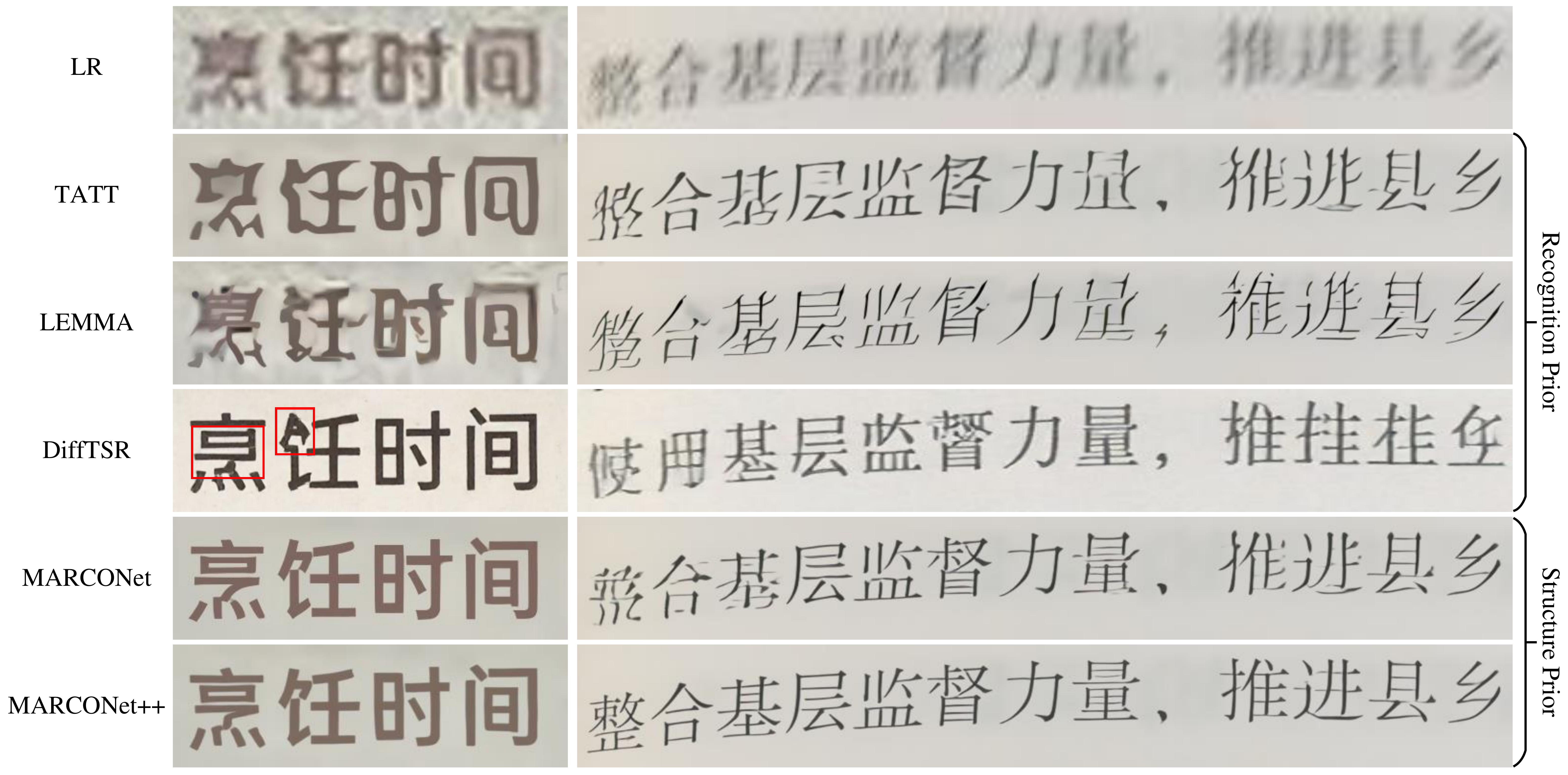}
    \caption{Results of applying recognition prior (TATT~\cite{ma2022text}, LEMMA~\cite{guo2023towards} and DiffTSR~\cite{zhang2023diffusion}) and structure prior (MARCONet~\cite{li2023marconet} and our enhanced MARCONet++) on real-world Chinese text images with regular and irregular layouts. TATT~\cite{ma2022text} and LEMMA~\cite{guo2023towards} are re-trained using our synthetic Chinese text images.}
    \label{fig:fig1}
    \vspace{-5pt}
\end{figure*}

{
To address the above challenges, existing methods primarily integrate high-level recognition priors into intermediate features~\cite{ma2021text,ma2022text,zhang2023diffusion,zhou2024recognition,zhao2023pean}.
While these approaches enhance recognition accuracy, they struggle to faithfully restore clear strokes from real-world LR input.
Figure~\ref{fig:fig1} presents two representative real-world LR inputs—one with a regular layout and the other with an irregular layout. TATT~\cite{ma2022text} and LEMMA~\cite{guo2023towards} are designed for English characters, whereas DiffTSR~\cite{zhang2023diffusion} targets real-world Chinese text. We observe that these recognition-prior-based methods struggle to generate accurate structures when the LR character contains complex strokes. The challenge intensifies when dealing with text images that have curved layouts (as shown in the right part of Figure~\ref{fig:fig1}). This suggests that high-level recognition priors provide only a rough constraint for the SR process, resulting in limited improvements in restoring accurate structures.
}

\begin{figure}[!t]
    \setlength{\abovecaptionskip}{2pt} 
    \setlength{\belowcaptionskip}{-12pt}
    \centering
    \includegraphics[width=.46\textwidth]{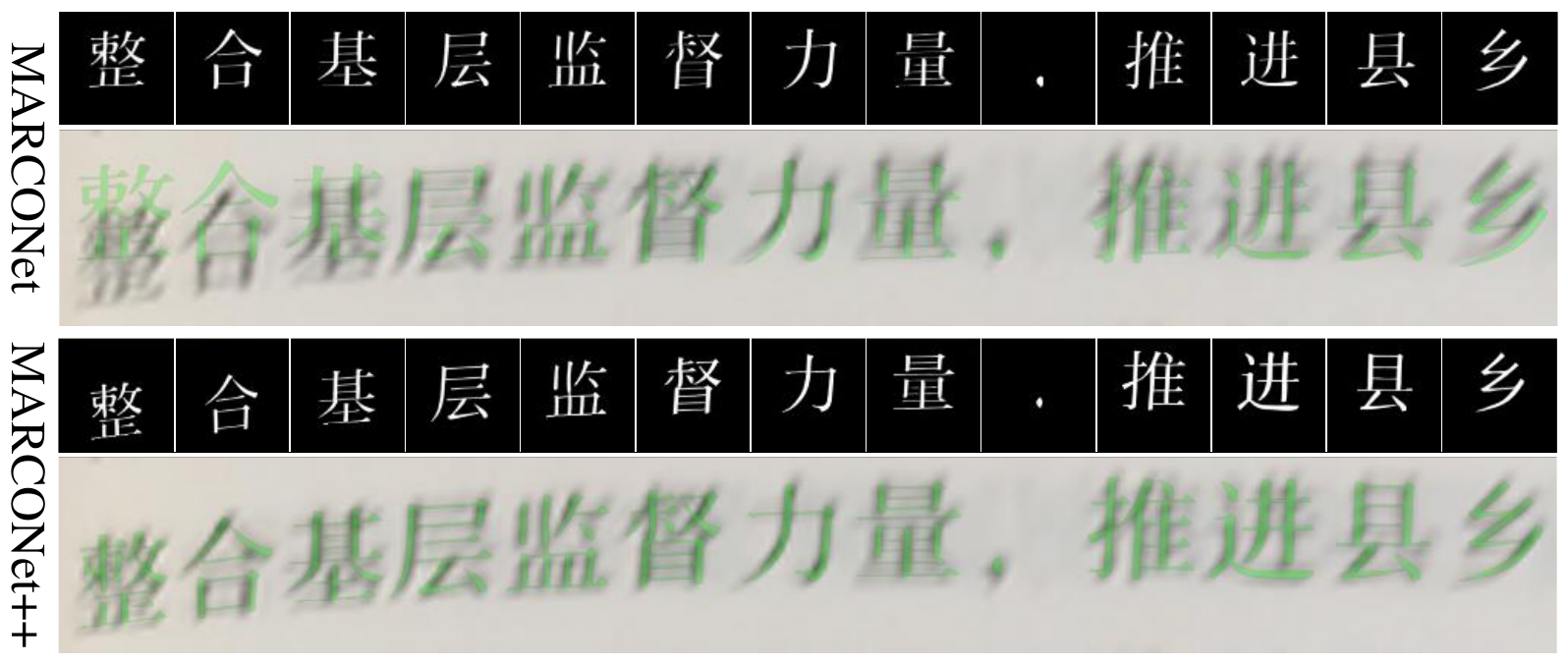}
    \caption{Visualization of the predicted structure prior (1-\textit{st} and 3-\textit{rd} rows) for MARCONet and MARCONet++ in Figure~\ref{fig:fig1}. The 2-\textit{nd} and 4-\textit{th} rows visualize the position of the structure prior (green part) on the LR text image.
    }
    \label{fig:fig1_ana}
    \vspace{-10pt}
\end{figure}

{
In this paper, we aim to explore the use of generative structure priors for restoring real-world low-resolution (LR) Chinese text images and to develop an effective method for adaptively incorporating these priors into various LR inputs.
To learn the structure prior for each character, we employ a StyleGAN-based generative model that produces structures consistent with LR characters. To ensure that the generated characters adhere to their specific structures while allowing for an almost infinite variety of styles, we reformulate the original StyleGAN by constraining the generative space within a codebook.
The codebook stores the discrete code of each character, with each code serving as a constant for StyleGAN to generate a specific high-resolution character. 
Figure~\ref{fig:fig1} shows that using structure prior can effectively restore accurate character structures. However, our previous model, MARCONet~\cite{li2023marconet}, performs limited on irregular layouts, as its vector $w$ controls only limited styles like typefaces. When applying MARCONet to irregular text images, the spatial misalignment between structure prior and LR characters can easily result in distorted structures (see the 1-\textit{st} row in Figure~\ref{fig:fig1_ana}).
To address this issue, we propose an enhanced model, MARCONet++, which learns a broader style space that includes typefaces, spatial arrangements, orientations, and perspectives. These diverse styles enable our structure prior to generalize effectively across real-world scenarios, which may involve different view perspectives or irregular arrangements (see the 2-\textit{nd} row in Figure~\ref{fig:fig1_ana}).
To apply the structure prior to each LR character, we employ two transformer-based encoder-decoder networks for predicting character locations and codebook indices.
With the style $w$, character location, and character code from the codebook, the structure prior is integrated into each LR character through the SR network.
}

{
This paper is a substantial extension of our previous work MARCONet~\cite{li2023marconet}. 
Compared to the conference version, we have enhanced MARCONet's capabilities to address a wider variety of general text SR scenarios.
1)~We expand the exploration of the character structure prior to accommodate a more general format, allowing for both regular and irregular text layouts. 
2)~We reformulate the $w$ prediction to capture more styles, including different typefaces, spatial locations, and orientations. 
We also restructure the jointly trained transformer encoders into three separate networks, without compromising model efficiency.
This adjustment allows for independent training of each model, which alleviates the challenges of balancing multiple tasks faced by the shared backbone in our previous MARCONet. As a result, this design leads to improvements of 12.0\% in LR character recognition accuracy and 41.2\% in location prediction accuracy.
Furthermore, this design enables the use of character classification models tailored to specific scenarios~\cite{chen2021benchmarking}, such as document images, scene images, and license plates, thereby improving character recognition accuracy.
3)~We enhance the pipeline of synthesizing training text images to cover a broader range of real-world scenarios, especially those involving irregular arrangements.
4)~We provide additional analyses and evaluations for more general cases. Experiments demonstrate that our method can effectively generate reliable and consistent structure prior
and generalize to other types of scripts, such as English, with competitive performance.
The enhanced structure prior facilitates the faithful restoration of detailed strokes and exhibits exceptional generalization abilities when applied to real-world LR text images.
}

We refer to this enhanced approach as {MARCONet++}. The main contributions are summarized as follows:
\begin{itemize}
    \setlength{\itemsep}{-2pt}
    \setlength{\parskip}{2pt} 
    \item We show that blind text SR tasks, especially for characters with complex structures and irregular layouts, can be faithfully restored by leveraging their structure prior encapsulated in a generative network.
    \item To learn the generative structure prior for each character, we reformulate StyleGAN by replacing its single constant with discrete codes. Each code represents a specific character and drives the StyleGAN to generate this character with diverse styles.
    \item We propose a practical framework for embedding the structure prior into LR text images by predicting character styles, locations, and their indexes within the codebook.
\end{itemize}

The remainder of this paper is organized as follows. In Section~\ref{sec:ref}, relevant studies including blind image SR, English and Chinese text image SR, and generative structural prior in image SR are reviewed. Section~\ref{sec:sec3} presents the details of learning the character structure prior and using it into text image SR. Section~\ref{sec:sec4} reports the experimental results and analyses. Finally, concluding remarks are provided in Section~\ref{sec:sec5}.

\section{Related Work}
\label{sec:ref}

%

\subsection{Blind Image Super-resolution}
Blind image super-resolution presents a significant challenge due to the complex mixture of unknown degradations. However, it is crucial for addressing real-world LR image restoration scenarios. 
Recent studies address the problem from three key aspects, \ie, estimating degradation parameters~\cite{bell2019blind,gu2019blind,luo2020unfolding,wang2021unsupervised,liang2022efficient}, establishing more realistic training data~\cite{cai2019toward,wei2020component,ji2020real,zhang2021designing,wang2021realesrgan,li2022from,yin2023metaf2n}, and designing efficient and effective networks~\cite{wang2023omni,zhou2023srformer,liang2021swinir,chen2022real,chen2023activating}. Among them, the first paradigm focuses on estimating degradation parameters of the degradation process and then applies non-blind SR methods, \eg, ZSSR~\cite{shocher2018zero}. The second category builds training pairs either through capturing real-world LR and HR pairs~\cite{cai2019toward,wei2020component} or designing elaborate degradation models that simulate real-world degradation~\cite{ji2020real,zhang2021designing,wang2021realesrgan,li2022from}. 
Nevertheless, when dealing with text images, which exhibit specific and semantic structures, we have shown that achieving high-quality restoration performance cannot rely solely on elaborately designed degradation models. The preservation of precise strokes in text SR is often of greater significance but receives relatively less attention.

\subsection{Scene Text Image SR}

Existing methods for scene text image SR have predominantly focused on English text images for many years, and this area holds significant value in real-world applications~\cite{dong2015boosting,xu2017learning,mou2020plugnet,wang2020scene,quan2020collaborative,chen2021scene,guo2023towards,nakaune2021skeleton,zhao2021scene,qin2022scene,zhao2022c3,chen2022text,ma2022text,zhu2023improving,zhao2023pean,zhou2024recognition}. 
In traditional methods, maximum a posterior (MAP)~\cite{capel2000super} and Bayesian framework~\cite{dalley2004single} are exploited for super-resolving text images. These earlier approaches have been unable to generate high-quality results.
Dong~\etal~\cite{dong2015boosting} employ CNNs~\cite{dong2015image} for text image SR and achieve promising results in the ICDAR 2015 competition~\cite{peyrard2015icdar2015}.
Xu~\etal~\cite{xu2017learning} adopt a Generative Adversarial Network (GAN)~\cite{goodfellow2014generative} to learn category-specific prior for face and text images SR, along with the supervision from a multi-class GAN loss.
Mou~\etal~\cite{mou2020plugnet} propose integrating a SR unit into the recognition process for degraded text images.
Wang~\etal~\cite{wang2020scene} introduce the first real-world scene text SR pairs (\ie, TextZoom), which are cropped from RealSR~\cite{cai2019toward} and SRRAW~\cite{zhang2019zoom}.
Both RealSR and SRRAW are captured by different digital cameras with different focal lengths, aiming to collect natural LR/HR pairs in real-world scenarios. They, along with Zhao~\etal~\cite{zhao2021scene}, present a sequential residual block by incorporating a bidirectional LSTM to capture the sequential information for low-level reconstruction.
We observe that most HR text images in this dataset are limited in clarity, which could limit the SR performance when taking them as ground-truth for learning the high-quality details.

{
Text SR could benefit from prior knowledge and auxiliary constraints. Several approaches have been proposed to improve the quality of text SR results.
Quan~\etal~\cite{quan2020collaborative} recover text images in a cascade model by predicting the high-frequency information. 
Chen~\etal~\cite{chen2021scene} propose Transformer-based position-aware and content-aware modules to emphasize the position and the content of each character.
LEMMA~\cite{guo2023towards} is another work that explicitly models character regions to produce high-level text-specific guidance for text SR.
Similarly, Nakaune~\etal~\cite{nakaune2021skeleton} and Qin~\etal~\cite{qin2022scene} introduce structure-aware loss and content-perceptual constraints, respectively, to learn detailed structural skeletons.
Zhao~\etal~\cite{zhao2022c3} propose C3-STISR by exploiting linguistic, recognition, and visual clues to jointly boost the SR performance.
Chen~\etal~\cite{chen2022text} develop a stroke-aware framework by concentrating on stroke-level internal structures.
Ma~\etal~\cite{ma2022text} introduce text recognition prior to text reconstruction with a Transformer-based module, leveraging the global attention mechanism.
They also embed categorical text priors in the encoder and employ multi-stage refinement to progressively enhance LR text images~\cite{ma2021text}. 
Guo~\etal~\cite{guo2023one} propose a one-stage framework that extracts multilevel knowledge from high-resolution images and transfers
it to the recognizer.
Zhao~\etal~\cite{zhao2023stirer} propose STIRER to effectively and simultaneously recover and recognize LR scene text images under a unified framework.
Zhu~\etal~\cite{zhu2023improving} propose an interesting Dual Prior Modulation Network (DPMN) to address the incorrect prior guidance caused by poor imaging conditions. Zhao~\etal~\cite{zhao2023pean}, Zhou~\etal\cite{zhou2024recognition}, Noguchi~\etal~\cite{noguchi2023scene} and Shrey~\etal~\cite{singhdcdm} propose to employ recognition prior in diffusion model to provide better guidance for scene text SR.
}

{
Most of the aforementioned methods incorporate recognition prior, either by employing it as a loss function on the SR results~\cite{chen2021scene,chen2022text,qin2022scene,zhao2022c3} or by leveraging it as intermediate SR features to provide high-level guidance~\cite{ma2021text,ma2022text,zhang2023diffusion,zhou2024recognition,zhao2023pean}. 
Although recognition prior is effective in improving text recognition, it is limited in providing accurate structure and style guidance, especially for certain texts with complex structures and various layouts. In this study, we demonstrate that generative structure prior offers better guidance for achieving more faithful restoration of character structures.
}

\subsection{Chinese Text Image SR}
{
Compared to English, Chinese characters exhibit more complex structures, making their faithful restoration more challenging and less explored. To the best of our knowledge, MARCONet~\cite{li2023marconet} is the first method to embed generative structure prior for the restoration of complex Chinese text.
Then, Ma~\etal~\cite{ma2023benchmark} propose the representative real-world Chinese-English scene text benchmark.
Recently, Zhang~\etal~\cite{zhang2023diffusion} present DiffTSR, which adopts text recognition features as a condition for the diffusion model for Chinese text SR. However, even with recognition prior in diffusion models, it still struggles to provide precise stroke-level guidance to achieve faithful restoration (see Figure~\ref{fig:fig1}). In comparison, our MARCONet++ provides accurate structure prior aligned in both spatial and structural dimensions, facilitating more faithful and clearer character restoration.
}

\subsection{Generative Structural Prior in Image SR}
The effectiveness of image structure prior has been demonstrated in many low-level vision tasks, \eg, depth image enhancement~\cite{li2016deep,hui2016depth,gu2017learning}, image inpainting~\cite{dolhansky2018eye,nazeri2019edgeconnect,ren2019structureflow}, and image restoration~\cite{pan2014deblurring,li2018learning,dogan2019exemplar,li2020enhanced,li2022learning}.
Most recently, by using generative structure priors obtained from pre-trained StyleGANs~\cite{karras2019style,karras2020analyzing}, codebooks~\cite{esser2021taming} or Stable Diffusion~\cite{rombach2022highresolution}, blind face restoration has achieved tremendous improvement~\cite{wang2021towards,chan2021glean,yang2021gan,chen2021progressive,li2020blind,gu2022vqfr,wang2022restoreformer,yue2022difface}, suggesting the apparent advantage of such prior over other methods in generating photo-realistic textures.
draws inspiration from these successful approaches.
However, crafting a suitable structural prior for text images presents a more complex challenge than face images. 
This is because each character has its unique strokes, yet may exhibit a vast variety of font styles and orientations. Any distorted, missing, or additional strokes easily change their semantic layout, leading to perceptible changes in their actual meaning (see Figure~\ref{fig:fig1}). 
All of these challenges aggravate the difficulties of using a generative structure prior to the text domain.
In this study, we design an effective way of learning such prior through replacing the constant input of StyleGAN with a discrete {codebook}, while controlling the font style $w$ via the $\mathcal{W}$ space. During the SR process, we predict the font style $w$ from LR input and then generate detailed structure guidance that aligns well with the LR character in terms of typeface, font size, and spatial location.

\section{Methodology}
\label{sec:sec3}
A GAN model trained on a large number of images from the same category can effectively capture the generative structure prior for that category. Many studies have explored using such generative priors to restore photo-realistic details from LR inputs~\cite{menon2020pulse,gu2020image,chan2021glean,yang2021gan,wang2021towards,pan2021exploiting}. However, previous works mainly focus on applying these priors to face images, while their use in enhancing text super-resolution (SR) remains relatively unexplored.
In this paper, we attempt to 
e\textbf{M}bed enhanced gener\textbf{A}tive st\textbf{R}u\textbf{C}ture pri\textbf{O}r for more general text SR, which we refer to as MARCONet++.
The overview of the framework is depicted in Figure~\ref{fig:pipeline_b}. 
Given a LR text input, MARCONet++ first obtains the generative structure prior for each LR character. To achieve this, we employ two transformer-based networks: one predicts the center location of each character, while the other predicts the character's index in the pre-trained codebook. We then apply GAN inversion (i.e., using the pSp encoder~\cite{richardson2021encoding}) to estimate the $w$ vector for each LR character.
Finally, the text SR network uses the generative structure prior as guidance for restoration. In the following sections, we will first discuss how we obtain this generative structure prior for each character and then introduce how to utilize it for text SR.

\subsection{Pre-training of Generative Structure Prior}
\label{sec:sec3_1}

\begin{figure}[!t]
    \centering
    \includegraphics[width=.49\textwidth]{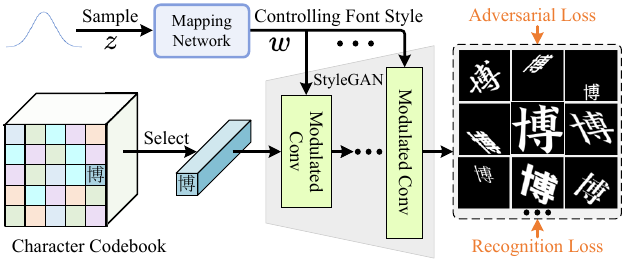}
    \caption{Pre-training of the enhanced generative structure prior for each character. 
    The codebook stores the discrete code of each character, and each code drives StyleGAN to generate a specific high-resolution character. Compared with MARCONet~\cite{li2023marconet}, the $w$ vector of this work controls not only the typefaces but also the font size, location, orientation, and perspective of each character (see the output).  The intermediate features encapsulate the generative structure prior and will be used for guiding text SR.
    }
    \label{fig:pipeline_a}
    \vspace{-10pt}
\end{figure}

The original StyleGAN2~\cite{karras2020analyzing} is designed to generate diverse, high-quality images within the same category. It takes a learnable constant as input and controls the image style through the {$w \in \mathcal{W}$ space}, which is a projection from Gaussian noise $z \in \mathcal{Z}$. Additionally, it introduces layer-wise noise to add stochastic variation for photo-realistic facial details.
To capture the unique structure of each character while still allowing for diverse styles, we remove the layer-wise noise and replace the single constant with discrete codes that represent different characters (see details in Figure~\ref{fig:pipeline_a}).
Each character code is denoted as $c\in \{C_i\}_{i=1}^M$, where $C$ is the {codebook} that stores each Chinese character. Each code is learnable and has a size of $1\!\times\!1\!\times\!512$.
Following CRNN~\cite{shi2016end}, the cardinality of codebook $M$ is set to 6,736.
The generated structure image $\mathcal{S}$ obtained through retrofitted StyleGAN is defined as: 
\begin{equation}
    \label{eqn:g}
    \mathcal{S} = G(c, \mathcal{F}(z);\Theta_{G}) = G(c, w;\Theta_{G})\,,
\end{equation}
where $\mathcal{F}$ is the network that maps $z$ to $w$, and $\Theta_{G}$ is the model parameters of StyleGAN.
For learning the character structure prior that focuse on structure only, we simplify the structure image $\mathcal{S}$ with pixel values $\in\!\{0,1\}$ (see the output in Figure~\ref{fig:pipeline_a}). 
Notably, the structure prior $\mathcal{P}$ is defined as the intermediate features from StyleGAN $G$ in Eqn.~(\ref{eqn:g}), which has the ability to reconstruct the corresponding high-quality character structures. The structure prior for character $c$ is defined as:
\begin{equation}
    \label{eqn:sp}
    \mathcal{P}^{c} = G_i(c, w;\Theta_{G})\,,
\end{equation}
where $G_i$ is the intermediate features from $i$\textit{-th} layer of $G$.

Since there are no high-resolution text images available for training our text StyleGAN, we propose synthesizing high-resolution character images using the PIL package\footnote{https://python-pillow.org/}. In particular, each HR character image $\mathcal{S}_\textit{GT}$ has the size of $128\!\times\!128$. During the synthesis process, we randomly set the font size, location, and typeface, and apply random perspective transformations for each character (see examples in Figure~\ref{fig:pipeline_a}). These augmentations help our text StyleGAN learn an enhanced structure prior with more general layouts, enabling it to generalize better to real-world text.
This represents a significant improvement over our previous work, MARCONet, which could only generate regular character images. The enhanced generative structure prior aligns well with real-world LR text images, which often exhibit varied and unconstrained conditions.
Additionally, unlike the original StyleGANs, which use only adversarial loss~\cite{goodfellow2014generative} to distinguish between real and fake images of faces, we introduce an additional recognition loss derived from a pre-trained character recognizer as regularization.
Once our model is trained, each learnable code $c$ in the codebook effectively stores the distinctive features of each character, while $w \in \mathcal{W}$ controls style-related attributes such as location, typeface, orientation, and perspective transformation.

\begin{figure*}[!t]
    \centering
    \includegraphics[width=.9\textwidth]{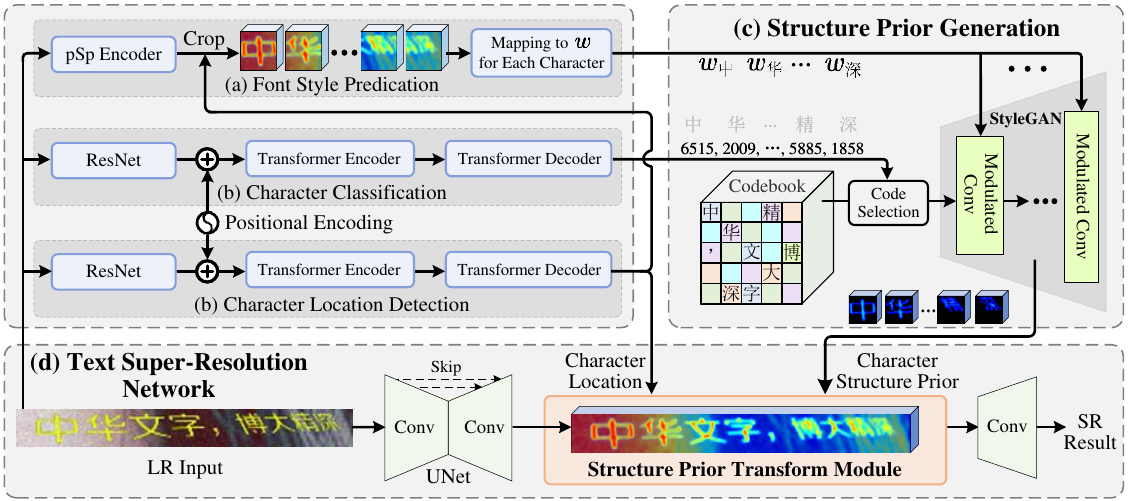}
    \caption{{Overview of our MARCONet++. It contains four parts. (a) A pSp encoder for obtaining the font style $w$ of each LR character. This enables the control of not only typefaces but also the spatial alignment between the structure prior and the LR character. (b) Two transformer-based encoder and decoder modules for predicting the character classification and location of each LR character. (c) Structure prior generation with a pre-trained StyleGAN for generating reliable structure prior for each character. (d) The SR process uses a structure prior transform module for embedding the structure prior into each LR character.}}
    \label{fig:pipeline_b}
    \vspace{-10pt}
\end{figure*}

\subsection{Style Prediction from LR Text Image}

An LR text image is usually composed of several characters, each of which may exhibit different styles when the text image is under perspective transformation or curved layout. For instance, characters on a curved layout might have varying locations (see Figure~\ref{fig:fig1_ana}).
{The previous MARCONet, which adopts a shared $w$ to capture typefaces for all characters, is not suitable for irregular scenarios. To address this limitation, we propose embedding an enhanced generative structure prior that accommodates more general layouts. Specifically, we predict the $w$ vector from the cropped character, allowing each character to have its own $w$ vector. This allows $w$ to control a wider range of styles (\eg, typefaces, orientations, and locations) for better alignment with each LR character.
}

Predicting the font style $w$ is similar to learning-based GAN inversion~\cite{tov2021designing,richardson2021encoding,wang2022high,alaluf2022hyperstyle}.
To this end, we adopt the pSp encoder~\cite{richardson2021encoding}, a well-known GAN inversion network,  to extract the features from the LR text image $I_\textit{LR}$. We then crop the character features based on the locations of each character and map them to $\mathcal{W}$ space using two linear layers. 
This style $w$ prediction branch can be optimized with the gradient from the StyleGAN inversion process. 
Specifically, given the pre-trained text StyleGAN $G$ and codebook $C$, the optimization of $w$ prediction network can be formulated as:
\begin{equation}
    \label{eqn:inv}
    \mathcal{L}_\textit{inv} = \sum_{i=1}^K \| G(c^i, w^i; \Theta_{w}) - S_\textit{GT}^i \|\,,
\end{equation}
where $c^i \in C$ represents the pre-trained code of $i$-th character in $I_\textit{LR}$ and $w^i$ is the predicted style obtained from the cropped feature of $i$-th character. $\Theta_{w}$ denotes the learnable parameters including the pSp encoder and the mapping to $w$. Here, $K$ is the number of characters in the LR text image, while $S_\textit{GT}^i$ is the corresponding ground-truth structure image for the $i$-th character. 
In this stage, we fix the parameters in the pre-trained text StyleGAN $G$ and codebook $C$ and optimize $\Theta_{w}$ only. 

We observe that two adjacent characters in the same text image usually exhibit very similar styles, \eg, font sizes and spatial locations. Hence, we further propose a regularization term to constrain the learning of $w$ prediction, defined as:
\begin{equation}
    \label{eqn:reg}
    \mathcal{L}_\textit{reg} = \sum_{i=1}^{K-1} \| w^{i+1} - w^{i} \|\,,
\end{equation}
where $w^{i+1}$ and $w^i$ represent font styles predicted from two adjacent characters, respectively.

The overall learning objective for predicting $w$ of each LR character is defined as:
\begin{equation}
    \label{eqn:w}
    \mathcal{L}_\textit{w} = \mathcal{L}_\textit{inv} + \lambda_\textit{reg} \cdot \mathcal{L}_\textit{reg}  \,.
\end{equation}

{
Once trained, our generated structure prior aligns the LR character with structural and spatial dimensions, effectively enhancing the text SR process for more general scenarios.
}

\subsection{Character Classification and Location Prediction}\label{sec:location}
During the SR process, another crucial aspect for generating structure prior is the code $c$ of each LR character. This code $c$ can be obtained by predicting the index in the codebook, which is a character classification task. 
To this end, we follow TransOCR~\cite{li2021trocr}, a Transformer-based encoder and decoder designed for character label prediction.  TransOCR~\cite{li2021trocr} is chosen so that we can better capture the dependency across different characters in the input image. 
It comprises a CNN backbone (\ie, ResNet45~\cite{he2016deep}), an image transformer as the encoder, and a text transformer as the decoder. 
Positional encoding is incorporated to inject positions of features in the sequence.
We employ cross-entropy loss for recognition, a widely used approach in text recognition tasks.  It's worth noting that our method is compatible with other text character recognition techniques or commercial models that may have superior accuracy in specific application scenarios, such as scene, web, document images, and license plates.

The generated structure prior has already aligned well with each LR character in both spatial and structural spaces. The next step is to predict the character location for injecting the structure prior into the LR input. To achieve this, we employ a similar transformer-based encoder and decoder for regressing the location. Since the predicted $w$ has already controlled the spatial location, here we only need to predict the center location along the width dimension of each LR character.
The ground-truth location of each character can be obtained when synthesizing the training images with the PIL package. 
The character location prediction framework can also be replaced with other superior works on specific application scenarios, without the need to fine-tune the StyleGAN and SR modules.

\subsection{Text Super-resolution Network} 
\label{subsec:network_architecture}

\begin{figure}[!t]
    \setlength{\abovecaptionskip}{5pt} 
    \setlength{\belowcaptionskip}{-12pt}
    \centering
    \includegraphics[width=.45\textwidth]{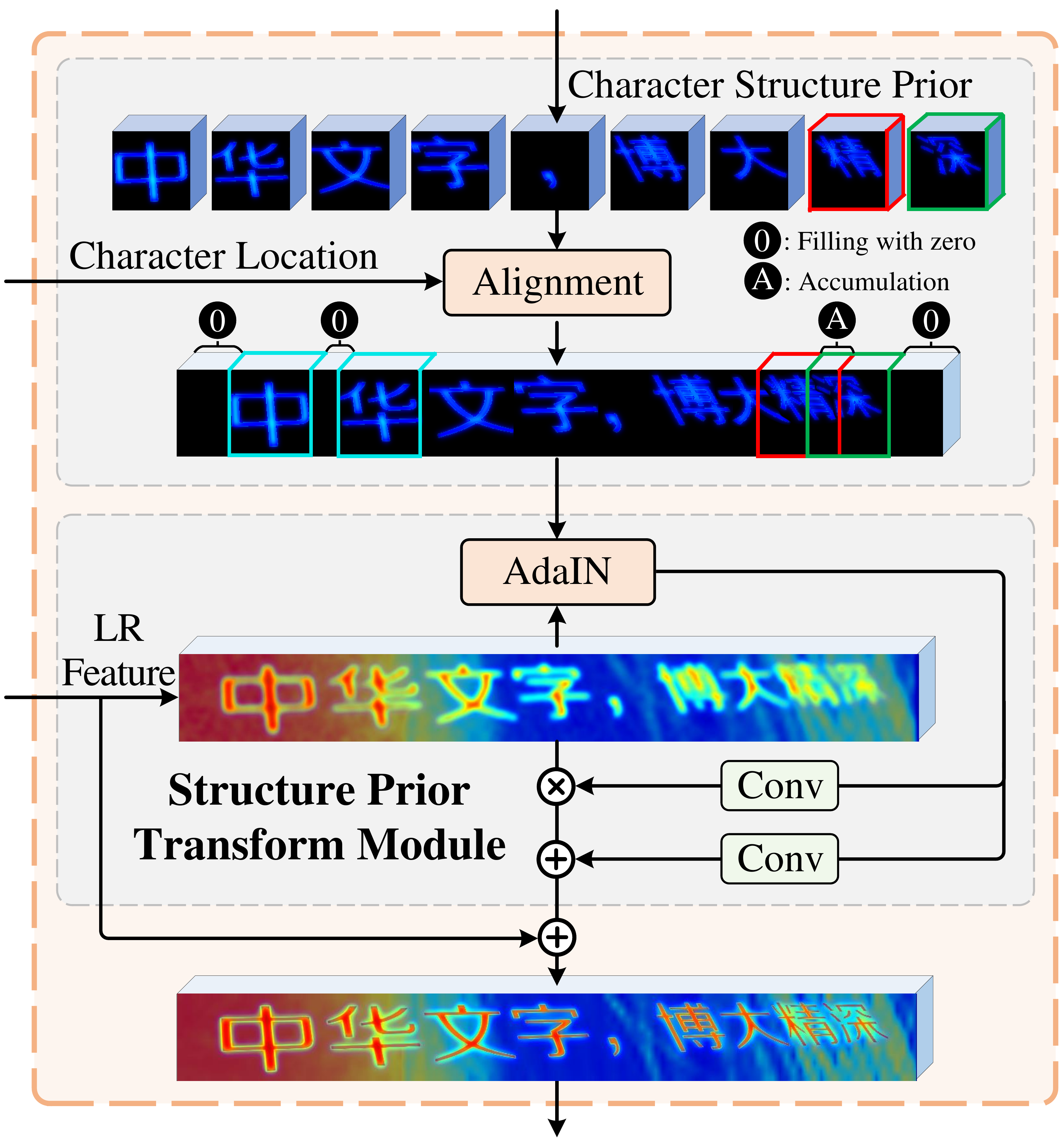}
    \caption{Details of the structure prior transform module. The structure prior is combined and aligned with the predicted character location. The LR text feature is then super-resolved with the guidance of their structure prior.}
    \label{fig:pipeline_c}
\end{figure}

\noindent\textbf{Network Architecture.} 
The aforementioned sub-networks predict the font style $w$, classification label (\ie, index in {codebook}), and the location of each character in the LR input. 
With them, the corresponding high-quality generative structure prior for LR each character can be generated through the pre-trained text StyleGAN using Eqn.~(\ref{eqn:sp}). 
Next, we describe how to use the structure prior for the text SR process. First, a simple UNet~\cite{ronneberger2015u} is adopted to extract the LR features. 
Then, the generative structure prior of each character is embedded into their LR counterparts through a structure prior transform module. The process is shown in Figure~\ref{fig:pipeline_c}.
Specifically, we align the prior with each LR character based on the detected character location and combine them to form a feature with the same size as the LR feature. Notably, the non-character region (\eg, the left space of character prior `\begin{CJK*}{UTF8}{gbsn}中\end{CJK*}') will be filled with zero. For the overlap region (\eg, the characters `\begin{CJK*}{UTF8}{gbsn}精\end{CJK*}' and `\begin{CJK*}{UTF8}{gbsn}深\end{CJK*}' marked with red and green boxes, respectively), we directly accumulate them because the feature visualization shows that the values in the background region of the structure prior are negligibly small.
Finally, AdaIN~\cite{huang2017arbitrary} is 
adopted to normalize the prior's distribution, and spatial feature transformation~\cite{wang2018recovering} is subsequently employed to predict the affine parameters. These scale and shift parameters are applied to the LR features. 
The structure prior transform module is adopted at two scales, \ie, with feature resolution of $s \in \{32, 64\}$, allowing our MARCONet++ to remain high fidelity with different degradation. 
A Conv-ReLU-ResBlock based CNN module is stacked to generate the final SR result.

\noindent\textbf{Learning Objective.} 
We minimize the differences between the SR results $\hat{I}_\textit{SR}$ and HR ground-truth $I_\textit{GT}$ on both the pixel and perceptual domains~\cite{johnson2016perceptual}:
\begin{equation}
    \label{eqn:rec}
    \mathcal{L}_\textit{rec}\!=\!\left\|\hat{I}_\textit{SR}\!-\!I_\textit{GT}\right\|_1 \!+\!\sum_{i=1}^4 \frac{\lambda_\textit{percep}}{\mathcal{C}_i\mathcal{H}_i\mathcal{W}_i}\left\|\Phi_i(\hat{I}_\textit{SR})\!-\!\Phi_i(I_\textit{GT})\right\|_1 \,,
\end{equation}
where $\mathcal{C}_i$, $\mathcal{H}_i$ and $\mathcal{W}_i$ are the feature dimensions from the $i$\textit{-th} convolution layer of the pre-trained VGG-19 model $\Phi$~\cite{simonyan2014very}. 
{$\lambda_\textit{per}$ is the trade-off parameter}. The loss~$\mathcal{L}_\textit{rec}$ is applied to the whole text image.

Adversarial loss~\cite{goodfellow2014generative} $\mathcal{L}_\textit{adv}$ is also {added} to improve visual quality. Instead of constraining on the whole image as in Eqn.~(\ref{eqn:rec}), $\mathcal{L}_\textit{adv}$ is performed on the cropped character image together with its corresponding structure image as additional conditions~\cite{mirza2014conditional}.
To be specific, the concatenation of HR character image $I^c_\textit{GT}$ and 
its structure image $\mathcal{S}^c_\textit{GT}$ is expected to be classified as Real, while the concatenation of SR character image $\hat{I}^c_\textit{SR}$ and its structure image $\mathcal{S}^c_\textit{SR}$ is recognized as Fake. Note that $\mathcal{S}^c_\textit{GT}$ is the ground-truth structure image of $I^c_\textit{GT}$, and $\mathcal{S}^c_\textit{SR}=G(c^i, w^i; \Theta_{w})$ is obtained from Eqn.~(\ref{eqn:g}), with $w^i$ and $c^i$ predicted from $i$-th LR character.
Such a design allows us to better constrain the embedding of generative structure prior into the SR results.
The adversarial loss~\cite{zhang2019self} is defined as:
\begin{gather}
    \scalebox{0.887}{$
        \begin{aligned}
            \mathcal{L}_D\!\!=\!\!-\mathbb{E}[\min(0,\!-\!1\!+\!D(I^c_{\textit{GT}}, \mathcal{S}^c_\textit{GT}))]\!-\!\mathbb{E}[\min (0,\!-\!1\!-\!D(\hat{I}^c_\textit{SR}, \mathcal{S}^c_\textit{SR}))]\,,\notag 
        \end{aligned}$} \\
    \scalebox{0.9}{$
        \mathcal{L}_G=-\mathbb{E}[D(\hat{I}^c_\textit{SR}, \mathcal{S}^c_\textit{SR})]\,.
        $}
\end{gather}

In the text SR training stage, we fine-tune StyleGAN, codebook, and $w$ prediction network using the $L_1$ constraint on the generated structure image $\mathcal{S}_\textit{SR}^c$. The learning rate is set to 1e-6. For each character $c$, this constraint is defined as:
\begin{equation}
    \setlength{\abovedisplayskip}{5pt}
    \setlength{\belowdisplayskip}{5pt}
    \mathcal{L}_\textit{str} = \|\mathcal{S}^c_\textit{SR}-\mathcal{S}^c_\textit{GT}\|_1\,.
\end{equation}

With these learning objectives, we can optimize the text SR branch and fine-tune the StyleGAN and codebook to enhance the generalization of the structure prior to the LR features. During this stage, the character classification and location prediction branches are kept fixed.
\begin{table*}[t]
    \centering
    \renewcommand\arraystretch{1.24}
    \caption{Quantitative comparison on \textbf{synthetic Chinese} text images. Note that all the competing methods are re-trained using the same data as ours. $\uparrow$ ($\downarrow$) indicates higher (lower) is better.}
    \setlength{\tabcolsep}{0.85mm}
    {
        \begin{tabular}{l| c c c c| c c c c| c c c c| c c c c}
            \hline
            \rowcolor{lightgray} & \multicolumn{8}{c|}{\textbf{Regular Layout}} & \multicolumn{8}{c}{\textbf{Irregular Layout}} \\
            \hhline{>{\arrayrulecolor{lightgray}}-|>{\arrayrulecolor{black}}----------------}
            \rowcolor{lightgray} & \multicolumn{4}{c|}{$\times2$} & \multicolumn{4}{c|}{$\times4$} & \multicolumn{4}{c|}{$\times2$} & \multicolumn{4}{c}{$\times4$}\\
            \rowcolor{lightgray}
            \multirow{-2.6}{*}{\makecell[c]{\textbf{Methods}}}&
            PSNR$\uparrow$ & SSIM$\uparrow$ &  LPIPS$\downarrow$ & ACC.$\uparrow$ & PSNR$\uparrow$ & SSIM$\uparrow$ &  LPIPS$\downarrow$ & ACC.$\uparrow$  & PSNR$\uparrow$ & SSIM$\uparrow$ &  LPIPS$\downarrow$ & ACC.$\uparrow$ & PSNR$\uparrow$ & SSIM$\uparrow$ &  LPIPS$\downarrow$ & ACC.$\uparrow$  \\
            \hline \hline
            SRCNN~\cite{dong2015image}   & 23.36 & .879 & .094 & 81.8 & 19.54 & .750 & .220 & 77.9 & 23.42 & .881 & .094 & 56.1 & 19.46 & .748 & .226 & 37.4\\
            ESRGAN~\cite{wang2018esrgan} & 23.71 & .891 & .090 & 82.1 &  19.84 & .777 & .209 & 78.1  & 23.68 & .890 & .086 & 56.1 & 19.83 & .762 & .214 & 37.6 \\
            Omni-SR~\cite{wang2023omni} & 24.12 & .886 & .082 & 85.5 & 20.57 & .780 & .193 &78.5 & 23.59 & .883 & .088 & 56.6 & 20.38 & .782 & .203 & 38.0  \\
            SRFormer~\cite{zhou2023srformer} & 24.66  & .894 & .082 & 86.3 &23.45 & .846 & .141 & 79.6 & 24.54 & .897 & .081 & 57.2 & 20.77 & .785 & .201 & 39.1 \\
            \hhline{-----------------}
            TSRN~\cite{wang2020scene}    & 24.25 & .892 & .096 & 87.2  & 20.64 & .780 & .212 & 81.2  & 23.98 & .891 & .097 & 58.3 & 20.36 & .780 & .219 & 39.4  \\
            TBSRN~\cite{chen2021scene}   & 25.60 & .905 & .076 & 88.4 & 21.02 & .793 & .191 & 81.5  & 25.13 & .902 & .076 & 58.6 & 20.95 & .792 & \underline{.194} & 39.2  \\
            TATT~\cite{ma2022text}       & 25.81 & .906 & .076 & 88.8  & 21.48 & .800 & .181 & 81.9 & 25.36 & \underline{.906} & .079 & 59.0  & \underline{21.30} & \underline{.796} & .197 & 39.5  \\
            LEMMA~\cite{guo2023towards} & 25.96 & .910 & .076 & 88.4 & 21.45 & .798 &.187 & 81.8& \underline{25.39} & \underline{.906} & .080 & \underline{59.2} & 21.29 & \underline{.796} &. 201 & \underline{39.7} \\
            MARCONet & \underline{28.21} & \underline{.932} & \underline{.041} & \underline{91.0} & \underline{23.96} & \underline{.867} & \underline{.098} & \underline{83.0} & 25.20 & .903 & \underline{.074} & 58.2 & 21.22 & .795 &.196 & 39.4 \\
            \textbf{MARCONet++}                & \bf{28.29} & \bf{.938} & \bf{.034} &  \bf{91.4} &  \bf{24.01} & \bf{.869} & \bf{.096} & \bf{83.2} & \bf{27.08} & \bf{.921} & \bf{.051} & \bf{60.1} & \bf{23.86} & \bf{.861} &\bf{.125}  & \bf{43.6} \\
            \hhline{-----------------}
            LR & - & - & - & 80.3 & - & - & - & 77.1 & - & - & - & 54.8 & - & - & - & 36.4 \\
            \hline
    \end{tabular}}
    \label{tab:com}
    \vspace{-6pt}
\end{table*}

\section{Experiments}
\label{sec:sec4}

\begin{figure*}[!t]
    \setlength{\abovecaptionskip}{3pt} 
    \centering
    \includegraphics[width=.9\textwidth]{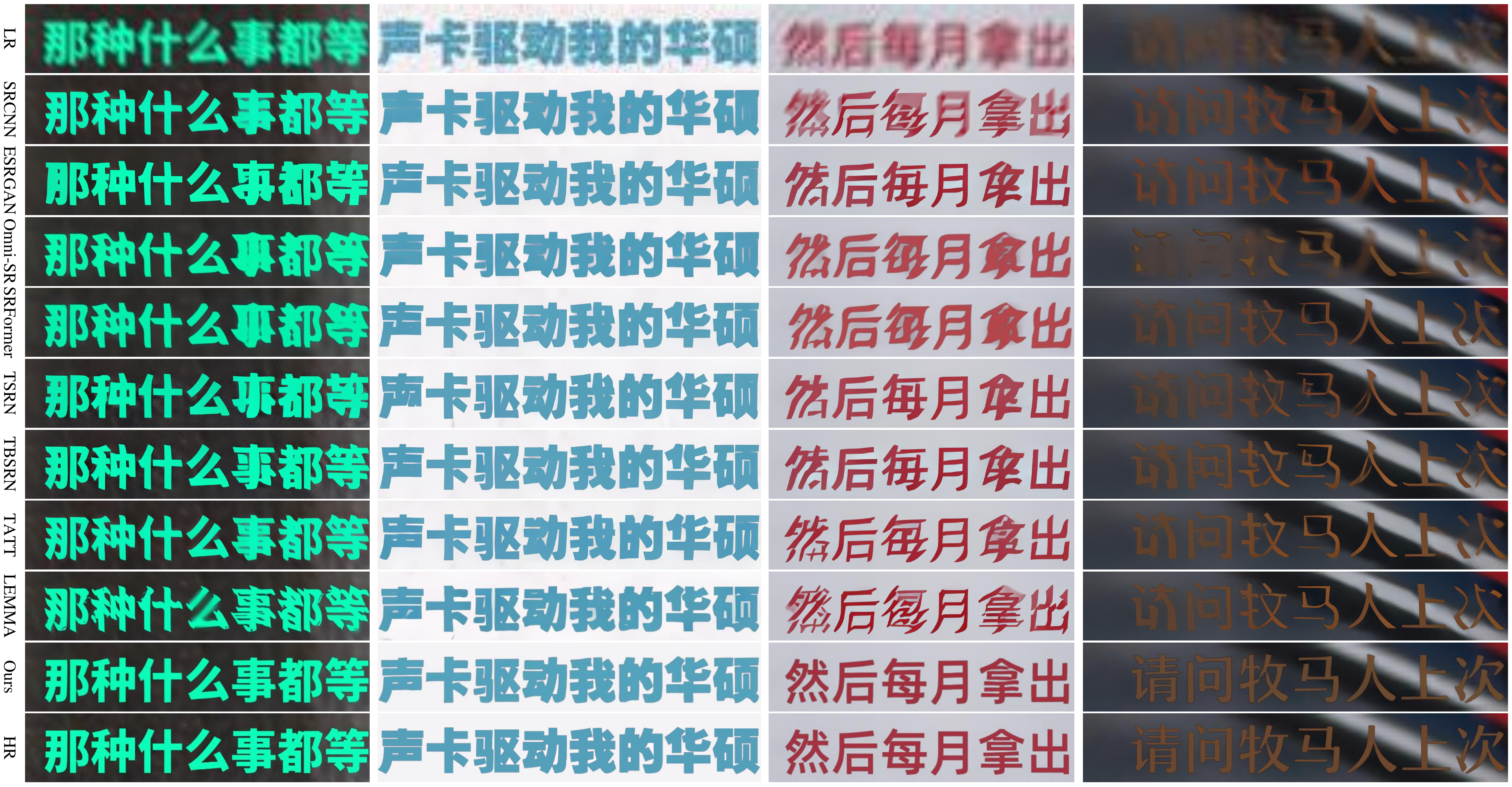}
    \caption{Visual comparison on $\times2$ ($1\!\sim\!2$ columns) and $\times4$ ($3\!\sim\!4$ columns) SR task under regular layout. The stroke details are best viewed by zooming in.}
    \label{fig:x2x4_regular}
    \vspace{-6pt}
\end{figure*}

\begin{figure*}[!t]
    \setlength{\abovecaptionskip}{3pt} 
    \centering
    \includegraphics[width=.9\textwidth]{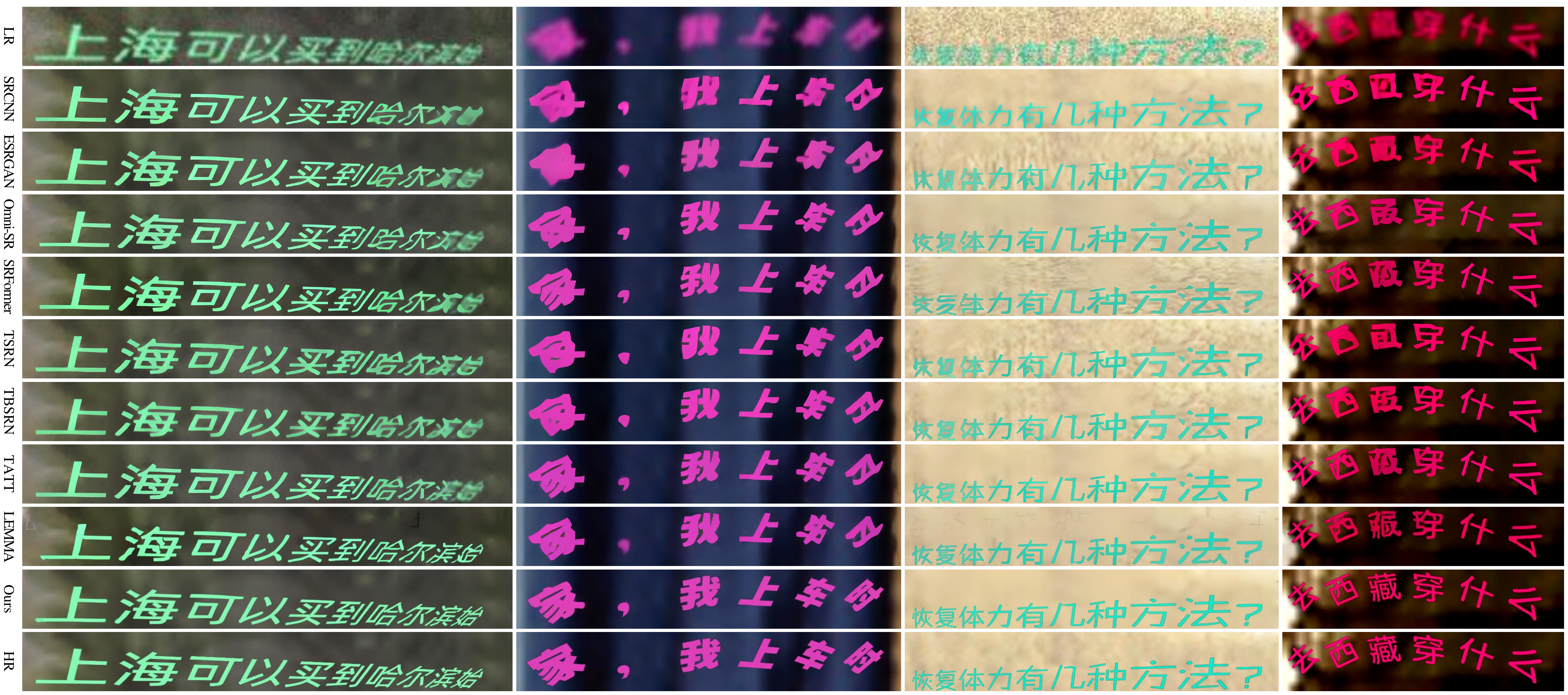}
    \caption{Visual comparison on $\times2$ ($1\!\sim\!2$ columns) and $\times4$ ($3\!\sim\!4$ columns) SR task under irregular layout. The stroke details are best viewed by zooming in.}
    \label{fig:x2x4_irregular}
    \vspace{-8pt}
\end{figure*}

\noindent\textbf{Training Data.}
{Although Ma~\etal~\cite{ma2023benchmark} propose the first Chinese-English text image benchmark, the LR and HR images contain multiple text lines. Additionally, there is no character location in this benchmark. These factors prevent our method from being trained effectively on it. We only evaluate the real-world SR performance on its test set.
To achieve faithful and high-quality Chinese text image SR, we propose a new pipeline capable of synthesizing high-quality text images with more general formats. Specifically, the synthetic images contain regular and irregular layouts. The regular layout is the same as in our previous work MARCONet (see examples in Figure~\ref{fig:x2x4_regular}). As for the irregular layout, we apply random perspective transformations to the regular and curved layouts (see examples in Figure~\ref{fig:x2x4_irregular}).
When synthesizing text images with curved distributions, the angle of each character is set within $\{-45^\circ, 45^\circ\}$, with the maximum height of the curve spanning up to two Chinese characters.
The PIL package is used to synthesize the regular text images, rendered with random RGB values, font sizes, and locations. The text is randomly selected from the Chinese corpus~\cite{bright_xu_2019_3402023}, which includes tens of millions of common items. Additionally, we collect 182 font families to introduce diverse font styles. 
The image background is obtained from the DIV2K~\cite{agustsson2017ntire} and Flick2K~\cite{timofte2017ntire} datasets, with each image randomly cropped and upsampled to $\times4\!\sim\!\!16$ times of the original size.
With the PIL toolbox, we can generate the HR text images $I_\textit{GT}$, together with the classification label, character location, and the ground-truth structure image $\mathcal{S}^c_\textit{GT}$ for each character.
The degradation pipelines presented in BSRGAN~\cite{zhang2021designing} and Real-ESRGAN~\cite{wang2021realesrgan} are applied online to degrade the HR image to LR input (see LR input in Figures~\ref{fig:x2x4_regular} and~\ref{fig:x2x4_irregular}).
This synthetic pipeline enables higher quality text restoration than using RealCE~\cite{ma2023benchmark} and TextZoom~\cite{wang2020scene}.
Real-world text images usually contain varying numbers of characters. Existing scene text SR methods typically resize the LR input to a fixed resolution for batch operation, potentially disrupting the intrinsic structure prior of each character. In contrast, we propose to pad with zeros along the width dimension to keep the structural ratio unchanged.
}

\noindent\textbf{Implementation Details.}
The whole training of our MARCONet++ is conducted on a server with four Tesla V100 GPUs. We employ Adam~\cite{kingma2014adam} as the optimizer. During the pre-training of the generative structure prior, the batch size is set to 16, with random perspective transformation on each structure image (see the examples in Figure~\ref{fig:pipeline_a}). When training the character classification and location prediction branches, the LR image is resized to $32\times256$ and the batch size is set to 48. 
As for the text SR branch, the batch size is set to 2. The height of HR images is set to 128, and the width after zero padding for SR is 2048. 
The synthetic text length contains 2$\sim$16 characters.
All the initial learning rates $lr$ are set to $2e{-4}$ and decreased by 0.5 when the losses reach a stable range on the validation set.
In the text SR training stage, the $lr$ for the StyleGAN and codebook is fixed to $1e{-6}$ for fine-tuning.
We also adopt color jittering~\cite{zoph2020learning} to increase image diversity. 
$\lambda_\textit{reg}$ and $\lambda_\textit{percep}$ are set to 0.02 and 0.05, respectively.
We first train the text StyleGAN for one day and then train the sub-networks for learning character classification and location for two days. Finally, it takes two days to train the text SR branch and fine-tune other networks.

\noindent\textbf{Baselines.}
{
To our knowledge, DiffTSR~\cite{zhang2023diffusion} is the only work that focuses on Chinese text SR. 
{While both MARCONet++ and DiffTSR address real-world Chinese text restoration, they differ fundamentally in their generative priors: MARCONet++ uses an explicit, structure-aware prior from a codebook and conditional StyleGAN, while DiffTSR relies on implicit semantic features in a diffusion process with limited structural guidance.}
In this work, we mainly compare it with our approach on real-world Chinese SR. We also report comparisons against general image SR methods (\ie, SRCNN~\cite{dong2015image}, ESRGAN~\cite{wang2018esrgan}, Omni-SR~\cite{wang2023omni}, and SRFormer~\cite{zhou2023srformer}) and English text image SR approaches (\ie, TSRN~\cite{wang2020scene}, TBSRN~\cite{chen2021scene}, TATT~\cite{ma2022text} and LEMMA~\cite{guo2023towards}).
There are other notable works on English text images SR, such as PEAN~\cite{zhao2023pean}, C3-STISR~\cite{zhao2022c3} and DPMN~\cite{zhu2023improving}. However, we did not reproduce them on Chinese text images because adapting their model designs to handle complex Chinese characters and training from scratch could easily affect model performance (as seen in the results of retrained TSRN, TBSRN, TATT, and LEMMA). For these competing methods that rely on text recognition priors in their English text SR tasks, we replace the recognizer with the corresponding pre-trained model for Chinese text images.
We report the quantitative and qualitative results on our synthetic regular and irregular layouts, with each group containing 1,000 LR\&HR pairs for $\times2$ and $\times4$ tasks, respectively. To simulate real-world degradation, these LR inputs are also injected with random noise, blurring, and JPEG compression. We also provide the visual results on real-world LR text images collected from the Internet and RealCE~\cite{ma2023benchmark}.
}

\subsection{Comparison on Synthetic Chinese Text Images}

{
Table~\ref{tab:com} presents the quantitative comparison of $\times2$ and $\times4$ text SR methods on both regular and irregular layouts. The evaluation metrics include PSNR, SSIM, and LPIPS~\cite{zhang2018unreasonable} from the IQA package~\cite{pyiqa}. The character recognition accuracy (ACC.) is obtained using SVTRv2~\cite{Du2022SVTR} from PaddleOCR.
We find that the general SR methods (\ie, SRCNN~\cite{dong2015image}, ESRGAN~\cite{wang2018esrgan}, Omni-SR~\cite{wang2023omni} and SRFormer~\cite{zhou2023srformer}) perform unsatisfactorily, as they are not specifically designed for text images. By incorporating the recognition information, TBSRN~\cite{chen2021scene}, TATT~\cite{ma2022text}, and LEMMA~\cite{guo2023towards} show significant improvements. Our previous MARCONet performs better on the regular layouts but fails on the irregular ones. 
In comparison, our proposed MARCONet++ achieves the highest scores across all scenarios, indicating superior performance in terms of image fidelity and similarity to the ground-truth. 
Additionally, our method exhibits the best recognition accuracy, suggesting faithful structure restoration in high-level space. 
}

\begin{figure*}[!t]
    \centering
    \includegraphics[width=.92\textwidth]{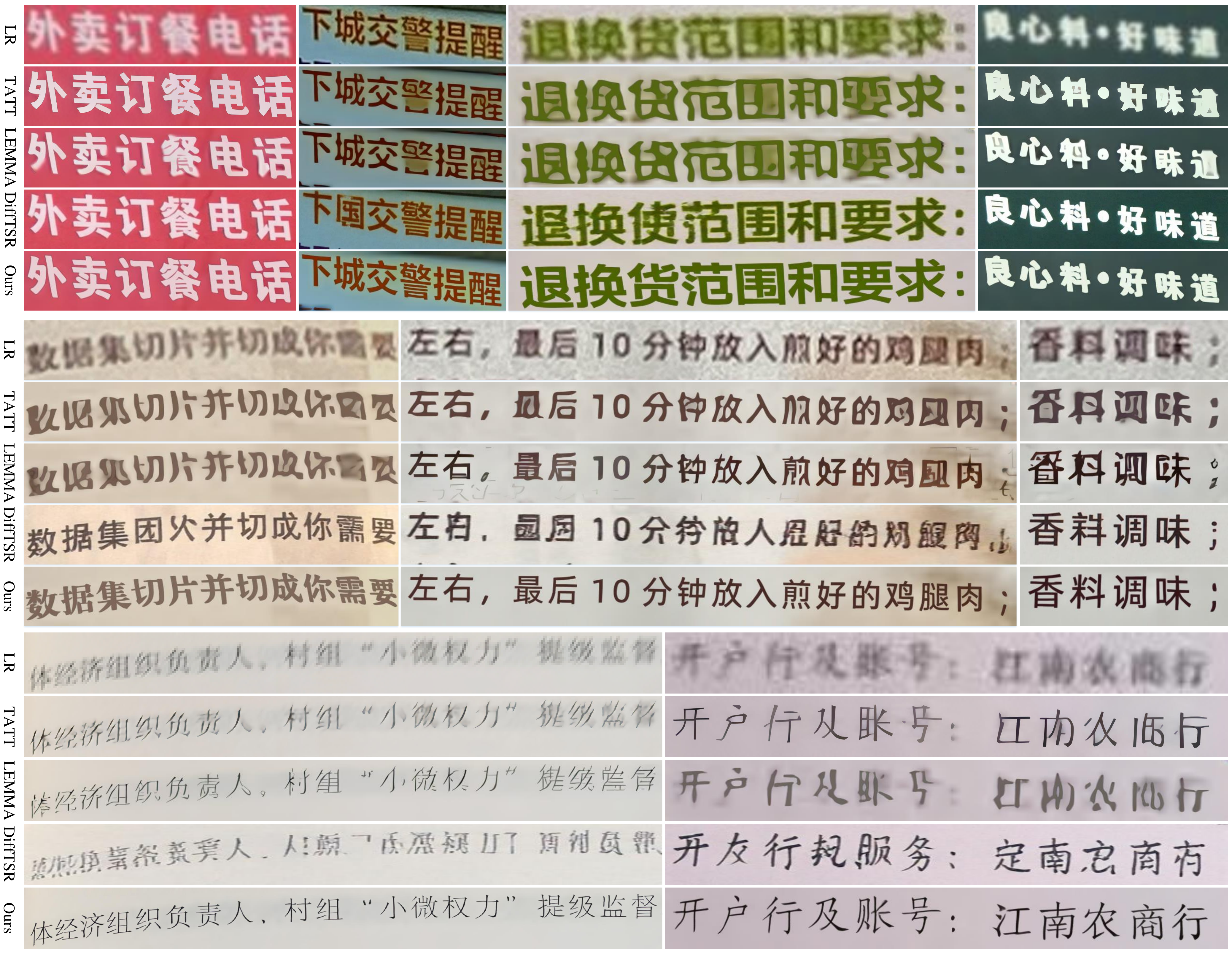}
    \caption{Visual comparison on real-world LR Chinese text segments covering different layouts. The stroke details are best viewed by zooming in.}
    \label{fig:real_segment}
\end{figure*}

Figures~\ref{fig:x2x4_regular} and~\ref{fig:x2x4_irregular} provide a visual comparison across different degradation levels and layout scenarios. When the LR input exhibits mild degradation, the competing methods demonstrate reasonable performance in restoring text details. However, they struggle to faithfully recover character structures with complex strokes, especially as the degradation worsens. This limitation is most evident in the last column, where all the competing methods fail to produce satisfactory results.
The results incorporating high-level recognition constraints offer limited benefits, particularly in handling diverse and complex font structures and styles. In contrast, our method, guided by the characters' generative structure prior, yields compelling results with more consistent structures aligned to the ground-truth.

\subsection{Comparison on Real-world Chinese Text Image}
\begin{figure*}[!t]
    \centering
    \includegraphics[width=.92\textwidth]{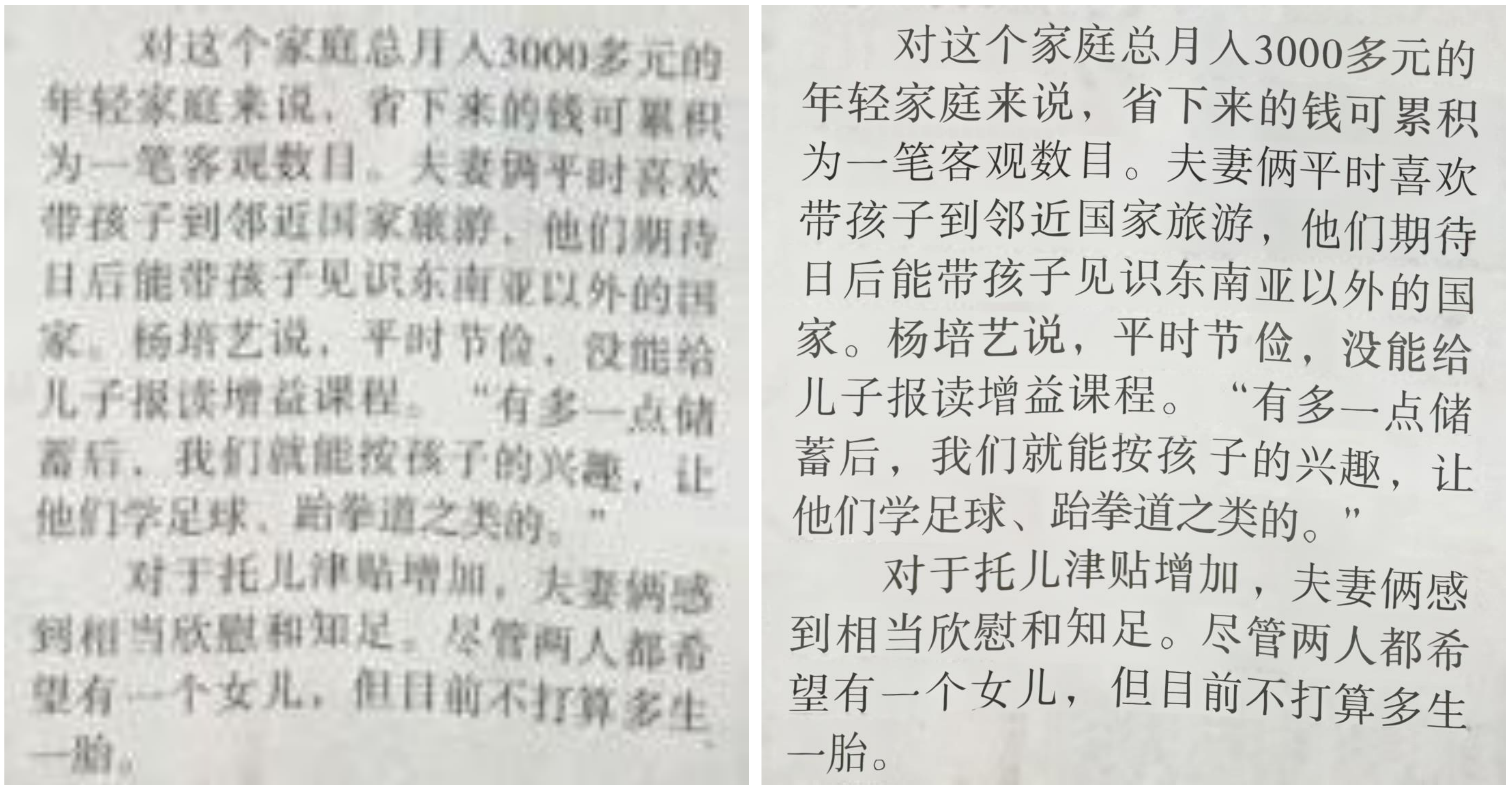}
    \caption{The restoration result on a real-world LR text region obtained from a newspaper. The stroke details are best viewed by zooming in.}
    \label{fig:real_whole}
\end{figure*}

{
Apart from the evaluation on synthetic LR input, we extend our assessment to real-world LR text images.
These LR images are collected from RealCE~\cite{ma2023benchmark} and the Internet, including captured invoices, documents, and scene texts, covering both regular and irregular layouts.
DiffTSR~\cite{zhang2023diffusion} is the only method specifically designed for real-world LR Chinese text images. Other methods are re-trained using our synthetic data.
Figure~\ref{fig:real_segment} shows that by leveraging our synthetic training data, the competing methods achieve comparable performance on characters with simple structures, even when dealing with LR inputs affected by unknown and complex degradation. 
However, these methods fail to generate faithful results when faced with severely degraded LR characters or those with more complex structures. 
Despite DiffTSR~\cite{zhang2023diffusion} using diffusion models, our MARCONet++ still outperforms it, not only in visual fidelity but also in accurately reconstructing character structures.
Figure~\ref{fig:real_whole} presents the restoration results of a real-world low-resolution newspaper. Notably, many lines in the text image show a visible curve, caused by folds during the newspaper's capture. Despite the challenges posed by varying paper quality and irregular layouts, our method consistently yields accurate and reliable results. 
}

\begin{table}[t]
    \centering
    \renewcommand\arraystretch{1.2}
    \setlength{\abovecaptionskip}{4pt}
    \caption{Evaluation on the \textbf{Real-world Chinese} test set. The best results are in \textbf{bold} and the second-best are \underline{underlined}.
    }
    \setlength{\tabcolsep}{0.2mm}
    {
        \begin{tabular}{l| c c c c| c c c c}
            \hline
            \rowcolor{lightgray} & \multicolumn{4}{c|}{\textbf{$\times2$}} & \multicolumn{4}{c}{\textbf{$\times4$}} \\
            \rowcolor{lightgray} \multirow{-1.4}{*}{\makecell[c]{\textbf{Methods}}}& PSNR$\uparrow$ & SSIM$\uparrow$ & LPIPS$\downarrow$ & ACC.$\uparrow$ & PSNR$\uparrow$ & SSIM$\uparrow$ & LPIPS$\downarrow$ & ACC.$\uparrow$ \\
            \hline \hline
            SRFormer & 17.40  & .6408  & .2816  & 74.8  & 16.62  & .6245  & .3131 & 64.9 \\
            TSRN & 17.42  & .6410  & .2801  & 76.5  & 16.66  & .6298  & .3087 & 66.3 \\
            TBSRN & 17.38  & .6406  & .2779  & 77.6  & 16.59  & .6217  & .3080 & 67.7 \\
            TATT & 17.42  & .6421  & .2769  & 78.4  & 16.65  & .6259  & .3094 & 69.0 \\
            LEMMA & 17.49  & .6441  & .2841  & 77.4  & 16.68  & .6300  & .3192 & 68.9 \\
            \hline
            DiffTSR & \underline{17.65}  & \underline{.6512}  & \bf{.2240}  & \underline{79.1}  & \underline{16.77}  & \underline{.6305}  & \bf{.2470} & \underline{69.4} \\
            MARCONet & 17.50  & .6443  & .3036  & 78.4  & 16.68  & .6304  & .3046 & 68.2 \\
            MARCONet++ & \bf{18.12}  & \bf{.6537}  & \underline{.2629}  & \bf{83.4}  & \bf{17.27}  & \bf{.6490}  & \underline{.2711} & \bf{76.4} \\
            \hline
            LR & - & - & - & 72.4 & - & - & - & 65.8 \\
            \hline
    \end{tabular}}
    \label{tab:realce}
    \vspace{-10pt}
\end{table}

{
To quantitatively evaluate real-world Chinese text images, we filter out images from the test set of RealCE that contain multiple lines and have inaccurate labels, constructing a new Chinese text benchmark, namely RealCE-1K. Table~\ref{tab:realce} demonstrates that although our MARCONet++ is trained on synthetic data, it shows superior performance compared to DiffTSR, indicating the effectiveness of our method in real-world scenarios. We will release RealCE-1K as a new benchmark in the future.
}

\subsection{Comparison on Real-world English Text Image}
\begin{figure}[t]
    \centering
    \includegraphics[width=.49\textwidth]{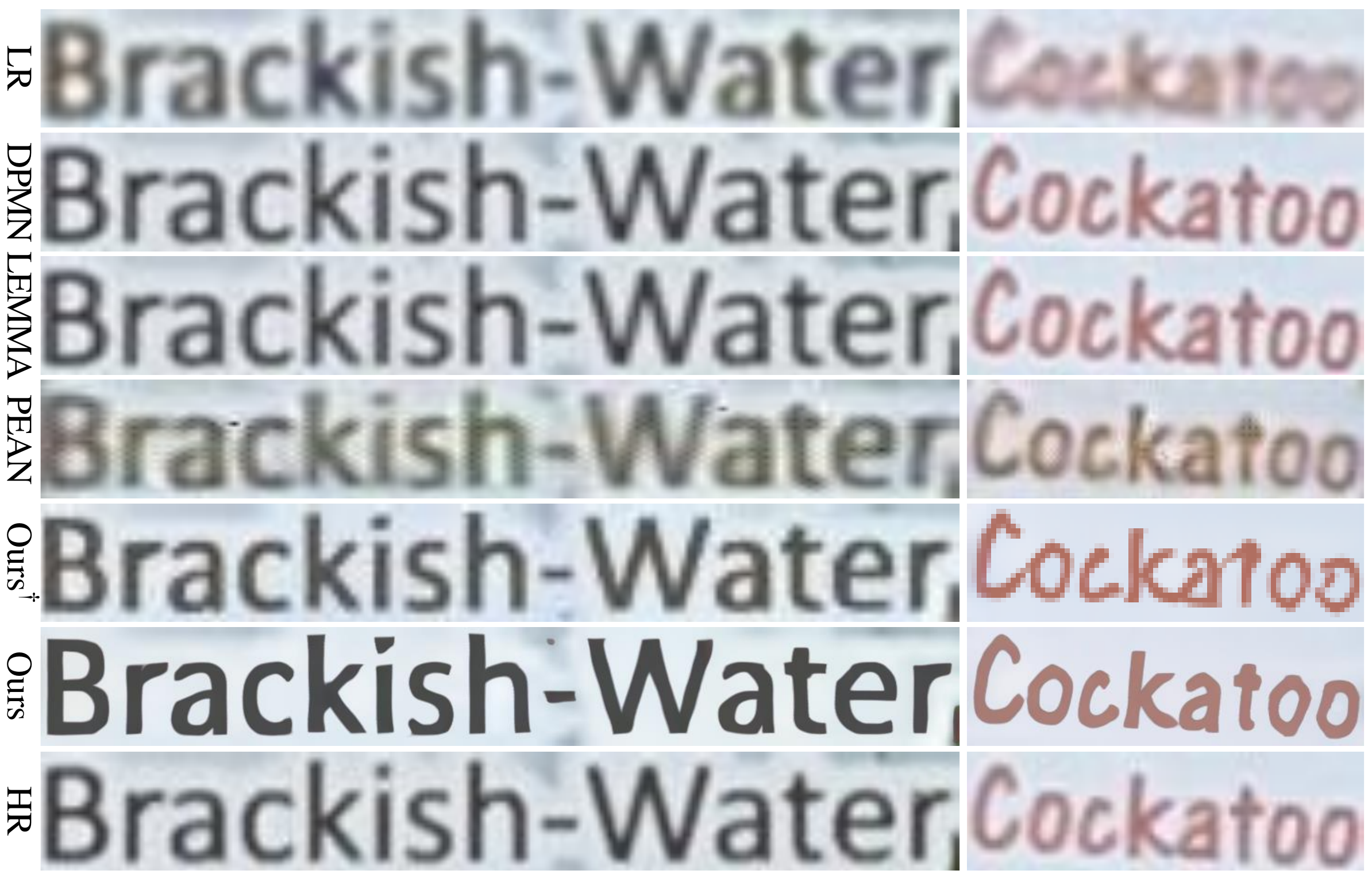}
    \caption{Visual comparison with English Text SR methods on TextZoom. Ours$^\dagger$ and Ours are trained on TextZoom and our synthetic data, respectively.}
    \label{fig:textzoom}
    \vspace{-8pt}
\end{figure}

{
TextZoom~\cite{wang2020scene} is a well-known dataset for scene-based English text SR. To evaluate on this dataset, we re-train our model using the provided LR\&HR pairs and character labels. Since TextZoom does not provide character locations, we use our pre-trained character location detection branch to obtain these locations. As shown in Table~\ref{tab:textzoom} and Figure~\ref{fig:textzoom}, 
our models struggle with this dataset. We find that the images in TextZoom lack clarity and that the detected character locations are not accurate for this dataset, both of which affect the mapping of the structural prior to the LR characters. We believe this issue could be addressed by collecting high-quality English text datasets with accurate character locations and labels.
}

\begin{table}[t]
    \centering
    \renewcommand\arraystretch{1.2}
    \caption{Evaluation on \textbf{Real-world English} text images (TextZoom). \\$^\ddagger$ indicates results are obtained using their released models.
    }
    \setlength{\tabcolsep}{3.9mm}
    {
        \begin{tabular}{l| c c c }
            \hline
            \rowcolor{lightgray} \textbf{Methods} & PSNR$\uparrow$ & SSIM$\uparrow$ & ACC.$\uparrow$ \\
            \hline \hline
            TATT~\cite{ma2022text} & 21.52 & .7930 & 63.6 \\
            C3-STISR~\cite{zhao2022c3} & 21.60 & .7931 & 64.1 \\
            DPMN (+TATT)~\cite{zhu2023improving} & 21.49 & .7925 & 63.9 \\
            LEMMA~\cite{guo2023towards}{$^\ddagger$} & 20.75 & .7727 & 66.0 \\
            RGDiffSR-1000\cite{zhou2024recognition} & \bf{21.88} & \bf{.7962} & 65.9 \\
            PEAN~\cite{zhao2023pean}$^\ddagger$ & 21.11 & .7491 & \bf{70.6} \\
            MARCONet++ (TextZoom) & 21.24 & .7526 & 62.9\\
            MARCONet++ (Synthetic) & 20.96 & .6724 & 54.2 \\
            \hline
    \end{tabular}}
    \label{tab:textzoom}
\end{table}

\begin{figure*}[!t]
    \centering
    \includegraphics[width=.9\textwidth]{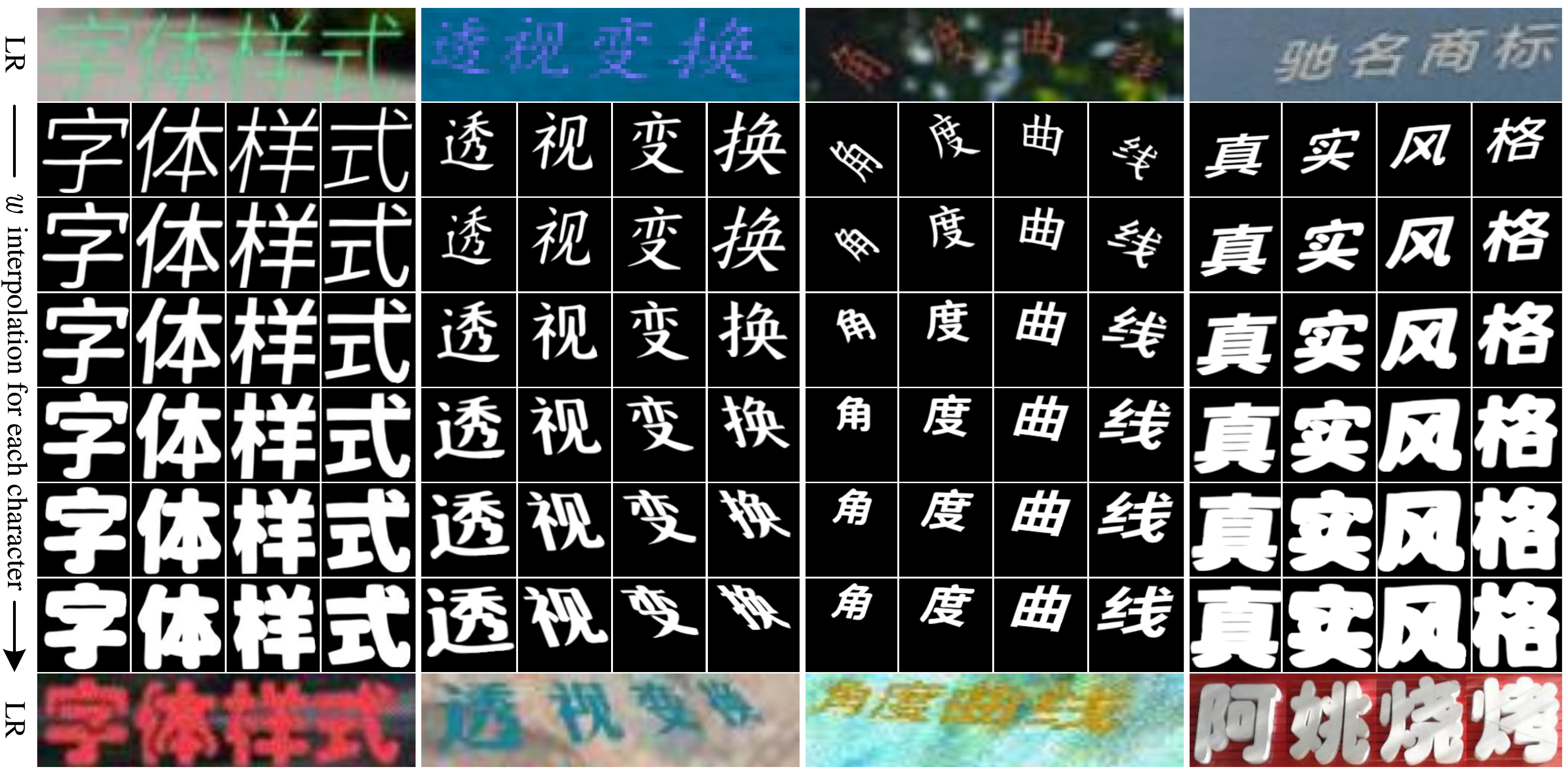}
    \caption{The structure image obtained with the interpolation of the $w$ vector obtained from two LR text images. It shows the $w$ vector in our text StyleGAN controls the font style, including typeface (1-\textit{st} column), perspective transformation (2-\textit{nd} column), character orientation (3-\textit{rd} column) and style transfer from real-world LR inputs. The 1-\textit{st} and 8-\textit{th} rows are the LR input. The 2-\textit{nd} and 7-\textit{th} rows are the output of StyleGAN with $w$ from each LR character.}
    \label{fig:w}
    \vspace{-8pt}
\end{figure*}

\subsection{Ablation Study}
\noindent\textbf{Analyses of $\mathcal{W}$ Space.} In this study, each code in the codebook defines the structure of a character, while the $\mathcal{W}$ space controls the stylistic aspects of the character's structure.  
Figure~\ref{fig:w} presents the structure image generated by our text StyleGAN using $w \in \mathcal{W}$ from each LR character.
It is evident that $w$ effectively captures various font styles, such as the thickness and italicization of different typefaces in the 1-\textit{st} column, perspective transformation in the 2-\textit{nd} column, character orientation in the 3-\textit{rd} column, and variations of font size and location in the 2$\sim$4-\textit{th} columns. 
Additionally, we demonstrate the interpolation of two $w$ vectors from different LR inputs in the $3\!\sim\!6$ rows. Results illustrate smooth transitions between generated structure images, indicating the editable ability of $w$ within our framework similar to the original StyleGAN.
Furthermore, we adopt the predicted $w$ from two real-world LR text images and explore font style transfer. Despite being trained on synthetic text images, our model exhibits remarkable adaptability in accurately capturing the styles of real-world text images. This capability enables successful style transfer to other characters, showcasing the model's robust generalization to diverse text layouts encountered in real-world scenarios.
Moreover, when adopting the $w$ vector obtained from the LR character `\begin{CJK*}{UTF8}{gbsn}驰\end{CJK*}' to another code $c=$`\begin{CJK*}{UTF8}{gbsn}真\end{CJK*}' (the 4-\textit{th} column), the generated structure image exhibits a similar style to `\begin{CJK*}{UTF8}{gbsn}驰\end{CJK*}' but has the same semantic structure as `\begin{CJK*}{UTF8}{gbsn}真\end{CJK*}'. This indicates a good disentanglement between $w$ and $c$ in controlling style and preserving the unique structure of each character, respectively.

\begin{table}[t]
    \centering
    \renewcommand\arraystretch{1.2}
    \caption{Comparison of different variants of our proposed method. 
    }
    \setlength{\tabcolsep}{1.34mm}
    {
        \begin{tabular}{l| c c c| c c c}
            \hline
            \rowcolor{lightgray} & \multicolumn{3}{c|}{$\times2$} & \multicolumn{3}{c}{$\times4$} \\
            \rowcolor{lightgray} \multirow{-2}{*}{\makecell[c]{\textbf{Variants}}}&
            PSNR$\uparrow$ & SSIM$\uparrow$ &  LPIPS$\downarrow$ & PSNR$\uparrow$ & SSIM$\uparrow$ &  LPIPS$\downarrow$ \\
            \hline \hline
            Ours (\textit{UNet})           & 23.54 & .883 & .093 & 19.68 & .751 & .213 \\
            Ours (\textit{UNet$^\dagger$}) & 23.63 & .889 & .091 & 19.89 & .763 & .209 \\
            Ours (\textit{w/o S})          & 25.38 & .910 & .063 & 21.45 & .799 & .163 \\
            Ours (\textit{w/o C})          & 25.15 & .908 & .069 & 21.20 & .793 & .179 \\
            Ours (\textit{\#32})           & 27.74 & .917 & .048 & 23.66 & .852 & .141 \\
            Ours (\textit{\#64})           & \underline{28.05} & \underline{.926} & \underline{.042} & \underline{23.81} & \underline{.859} & {.129} \\
            Ours (\textit{D$^\textit{--}$})& {27.87} & {.921} & \underline{.042} & 23.73 & .857 & \underline{.126} \\
            \textbf{Ours (\textit{Full})}  & \bf{28.08} & \bf{.927} & \bf{.041} & \bf{23.86} & \bf{.861} & \bf{.125} \\
            \hline
            $\lambda_\textit{percep}=0.0$  & \bf{28.10} & \bf{.928} & .058 & \bf{23.89} & \bf{.862} & .197 \\
            $\lambda_\textit{percep}=0.05$ & 28.08 & .927 & .041 & 23.86 & .861 & .125\\
            $\lambda_\textit{percep}=0.1$ & 27.88 & .921 & \bf{.039} & 23.71 & .856 & \bf{.122}\\
            $\lambda_\textit{reg}=0.0$ & 27.76 & .916 & .045 & 23.58 & .849 & .143\\
            $\lambda_\textit{reg}=0.02$ & 28.08 & .927 & .041 & 23.86 & .861 & .125\\
            $\lambda_\textit{reg}=0.04$ & 27.78 & .917 & .042 & 23.67 & .852 & .136 \\
            $\lambda_\textit{reg}=0.1$ & 27.70 & .914 & .042 & 23.61 & .850 & .141 \\
            \hline
    \end{tabular}}
    \label{tab:aba}
    \vspace{-8pt}
\end{table}

\begin{figure}[t]
    \centering
    \hspace{-8pt}
    \includegraphics[width=.48\textwidth]{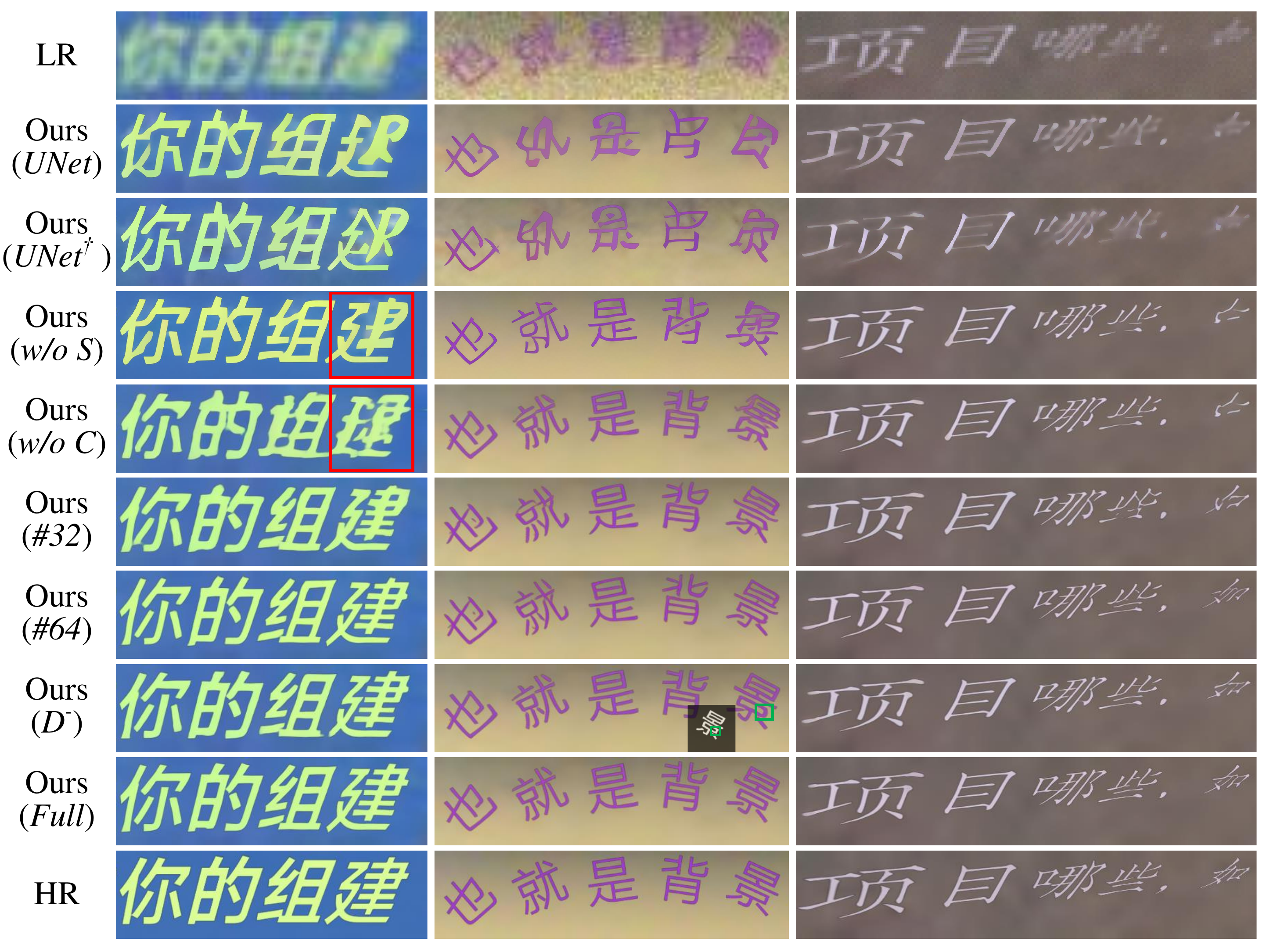}
    \caption{Visual comparison of different variants. Zoom in to see details.}
    \label{fig:aba}
    \vspace{-8pt}
\end{figure}

\noindent\textbf{Analyses of Variants.} Here we consider the following variants to evaluate each part of our MARCONet++, 
1) Ours~(\textit{UNet}) \& Ours~(\textit{UNet$^\dagger$}): only taking UNet to perform the SR task and increasing its model parameters, respectively, 
2)~Ours~(\textit{D$^\textit{--}$}): only taking the SR and HR images as input of discriminator without concatenating their structure images, 
3)~Ours~(\textit{w/o~S}): removing StyleGAN and directly incorporating the code $c$ on each LR character features through spatial affine transformation,
4)~Ours~(\textit{w/o~C}): removing {codebook} and directly concatenating the LR character features into the pre-trained StyleGAN like GFPGAN~\cite{wang2021towards},
5)~Ours~(\textit{\#32}) \& Ours~(\textit{\#64}): only using the structure prior transform module on feature sizes of $32\times32$ and $64\times64$.
Table~\ref{tab:aba} and Figure~\ref{fig:aba} show their performance on our synthetic irregular test set.
It can be observed that 1) both Ours~(\textit{UNet}) and Ours~(\textit{UNet$^\dagger$}) perform on par with the general SR methods. This suggests that simply increasing the model capacity does not bring significant performance improvements;
2) By removing StyleGAN and codebook, Ours~(\textit{w/o~S}) and Ours~(\textit{w/o~C}) show obvious inferior results, indicating that codebook can benefit the structure details while StyleGAN contributes more to the stroke quality (see the red box). The combination of them in Ours~(\textit{Full}) yields a performance enhancement greater than the sum of their individual contribution;
3) The {green box} in Figure~\ref{fig:aba} shows that without concatenating the structure image on the discriminator, the restored strokes have slight distortions compared to those in Ours~(\textit{Full}). This indicates that our discriminator is beneficial for emphasizing the structure prior more on the LR input and thus boosting the final SR result;
4) Ours~(\textit{\#64}) has slightly better results than Ours~(\textit{\#32}) but both of them are inferior to Ours~(\textit{Full}) that deploys a multi-scale structure prior transform module.
From these analyses, we conclude that the integration of the codebook and StyleGAN plays a crucial role in providing accurate structure guidance for the SR process. Additionally, the design of the multi-scale structure prior transform module and discriminator further benefits the performance. 

\begin{figure}[!t]
    \centering
    \includegraphics[width=.48\textwidth]{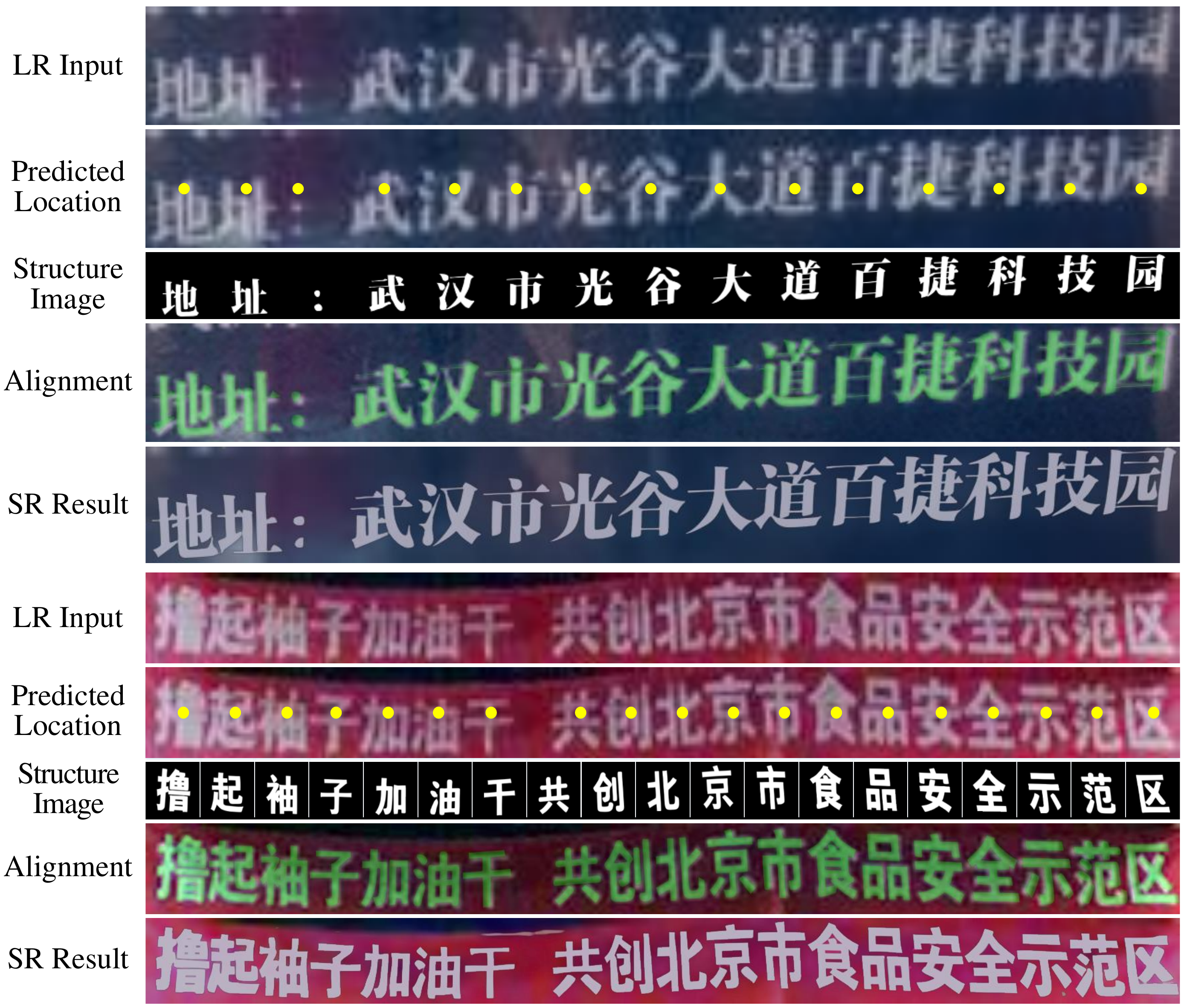}
    \caption{Visualization of detected character location (2-\textit{nd} row), generated structure image using $w$ and $c$ from LR character (3-\textit{rd} row), spatial alignment of the structure prior with LR character (4-\textit{th} row), and SR results (5-\textit{th} row).
    }
    \label{fig:vis}
\end{figure}

{
The results of using different trade-off parameters $\lambda_\textit{reg}$ and $\lambda_\textit{percep}$ in Eqns.~\ref{eqn:w} and \ref{eqn:rec} are shown at the bottom of Table~\ref{tab:aba}. One can see that perceptual loss is beneficial for LPIPS, while the regularization of $w$ boosts the SR results. A higher $\lambda_\textit{reg}$ has an adverse effect as it enforces the adjacent characters having the same styles, which is not applicable for irregular layouts.
}

\noindent\textbf{Visualizing Outputs of Major Modules.} 
{
Figure~\ref{fig:vis} visualizes the predicted character locations (the 2-\textit{nd} row) and structure images (the 3-\textit{rd} row) on two real-world LR segments. 
The generated structure image is obtained using $w$ and $c$ predicted from the LR character, which indicates that our $w$ can effectively capture the styles (\eg, typeface, location, and size) from the LR input. By using the character location, the visualization of embedding the structure prior into LR text is shown in the 4-\textit{th} row (the green structure is the predicted structure image). This suggests that the predicted character's location and style $w$ collaborate to align the structure prior with the LR characters.
}

\begin{table}[t]
    \centering
    \caption{Quantitative results for LR images with different text lengths. 
    }
    \setlength{\tabcolsep}{2.6mm}
    {
        \begin{tabular}{c| c c c c c}
            \hline
            \rowcolor{lightgray} \makecell[c]{\textbf{Text Length}}&
            PSNR$\uparrow$ & SSIM$\uparrow$ &  LPIPS$\downarrow$ & LOC.$\downarrow$ & ACC.$\uparrow$ \\
            \hline
            \hline
            4 & 24.86 & .872 & .0669 & 28.21 & 85.3 \\
            8 & 24.89 & .872 & .0656 & 28.06 & 87.1 \\
            16 & 24.48 & .860 & .0663 & 41.57 & 65.4 \\
            32 & 23.87 & .845 & .0669 & 91.83 & 48.1\\
            \hline
    \end{tabular}}
    \label{tab:diflen}
    \vspace{-6pt}
\end{table}

\noindent\textbf{Effect of Text Length.} 
{
To analyze the effect, we synthesize 1,000 text images, each containing 32 characters, and then crop each image into text segments of 4, 8, and 16 characters, respectively. From Table~\ref{tab:diflen}, we can observe that when the text length is around 8, both the predicted location (LOC.) and the character accuracy (ACC.) have minor errors compared to the ground-truth, leading to better image SR quality. However, when the text length exceeds 16, the metrics for LOC. and ACC. decline significantly, resulting in worse SR performance.
This is because, during the training stage, the input used for learning character location and recognition consists of a random number of characters (2$\sim$16). 
When the LR text image contains fewer than 16 characters, the difference in SR performance is negligible. 
For restoring LR inputs with longer sentences, we recommend cropping inputs to segments with fewer than 16 characters for improved text restoration.
}

\begin{figure}[!t]
    \setlength{\abovecaptionskip}{5pt} 
    \setlength{\belowcaptionskip}{-12pt}
    \centering
    \includegraphics[width=.49\textwidth]{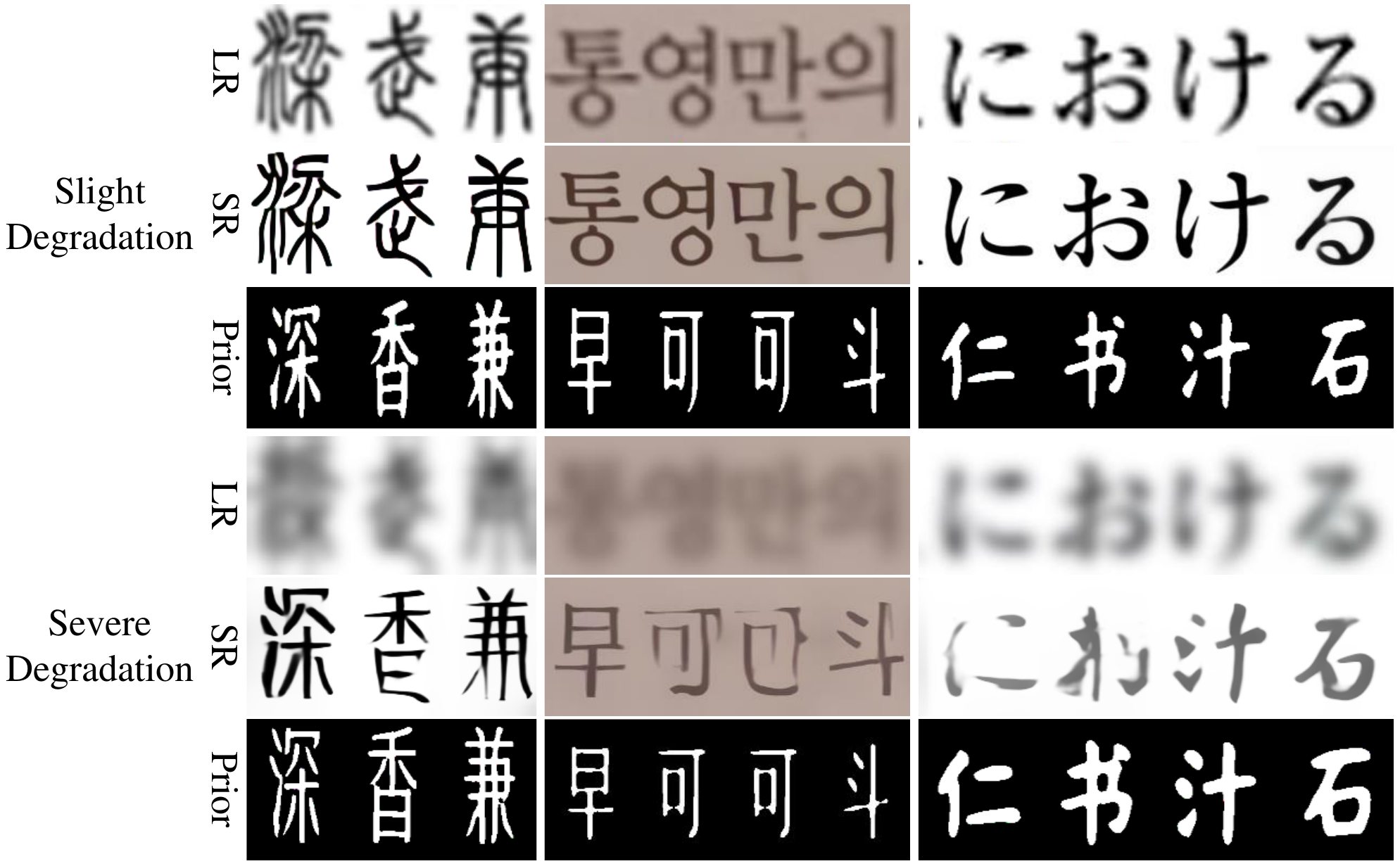}
    \caption{Analyses of generalization ability across different languages. From left to right: traditional Chinese, Korean, and Japanese text images. }
    \label{fig:otherlanguage}
    \vspace{-8pt}
\end{figure}

\noindent\textbf{Generalization to other Languages.}
{
Our model is specifically trained on Chinese text images. Figure~\ref{fig:otherlanguage} shows that when applied to other languages (i.e., traditional Chinese, Korean, and Japanese), our method can still perform favorably if the input is only slightly degraded. However, if the input experiences severe degradation, the restoration results are more consistent with the inaccurate structural prior. Therefore, it is preferable to train our model on these languages for improved performance.}

\noindent\textbf{Improvmeent of MARCONet++ Upon MARCONet.}
{
The most significant improvement of MARCONet++ over MARCONet is the enhanced structural prior, which can be applied to characters with different irregular layouts, allowing it to adapt to a wider range of real-world scenarios. Additionally, the separate training for predicting character location and recognition enables MARCONet++ to achieve higher accuracy (a 12.0\% improvement in ACC. and 41.2\% in LOC.). This is not merely a simple data modification but rather a more appropriate enhancement tailored to real-world conditions.
}

\noindent\textbf{Inference Time.} 
{
It takes 48.6 \textit{ms} for our method to restore the LR text images with an average size of $32 \times179.3$. Specifically, character location and recognition need 1.86 \textit{ms}, while $w$ prediction and the SR process take 9.28 \textit{ms} and 37.46 \textit{ms}, respectively. Our method is comparable to LEMMA~\cite{guo2023towards} (31.50 \textit{ms}) but is obviously faster than DiffTSR~\cite{zhang2023diffusion} (36090 \textit{ms}).
}

\begin{figure*}[!t]
    \centering
    \includegraphics[width=1\textwidth]{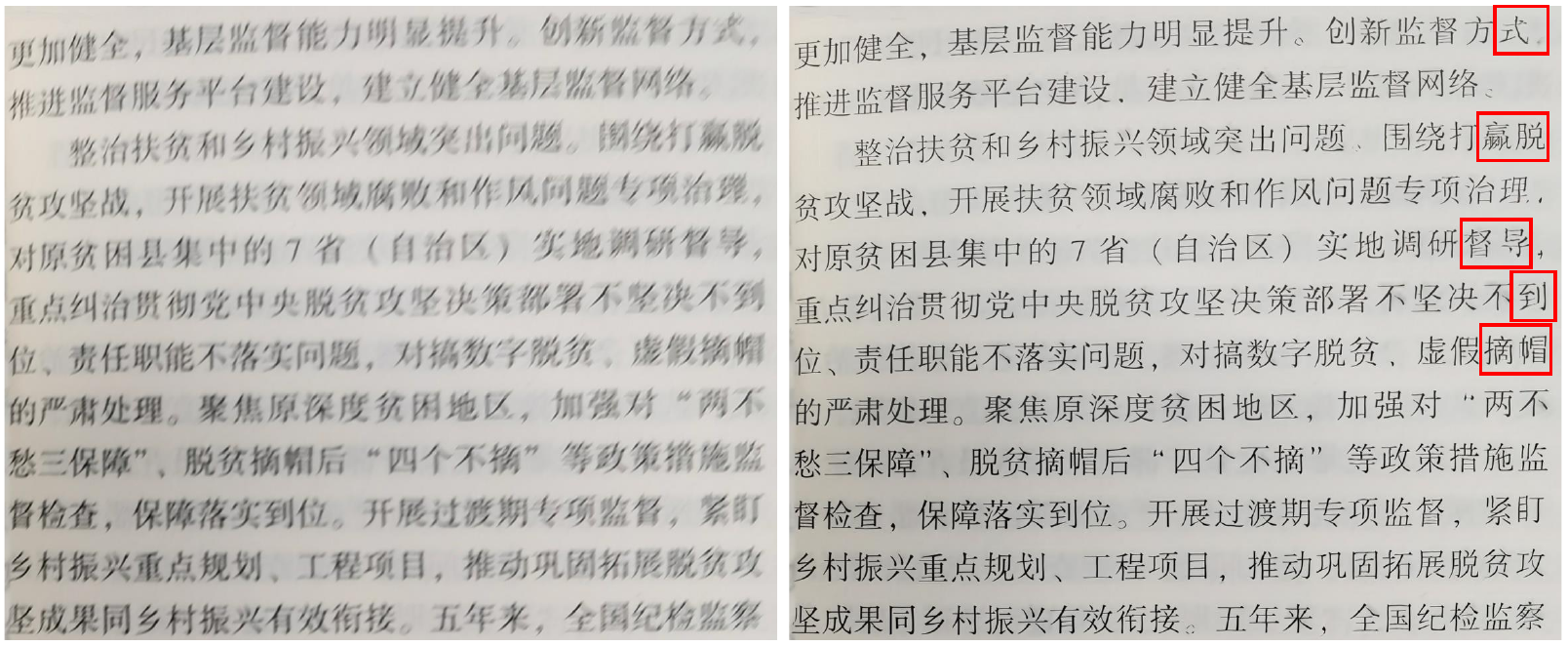}
    \caption{{Restoration results on real-world motion-blurred text images. Left: real captured inputs affected by motion blur. Right: outputs restored by our MARCONet++. Zoom in to observe the improved legibility and structural fidelity.}}
    \label{fig:motion}
\end{figure*}

\begin{figure}[!t]
    \centering
    \includegraphics[width=.48\textwidth]{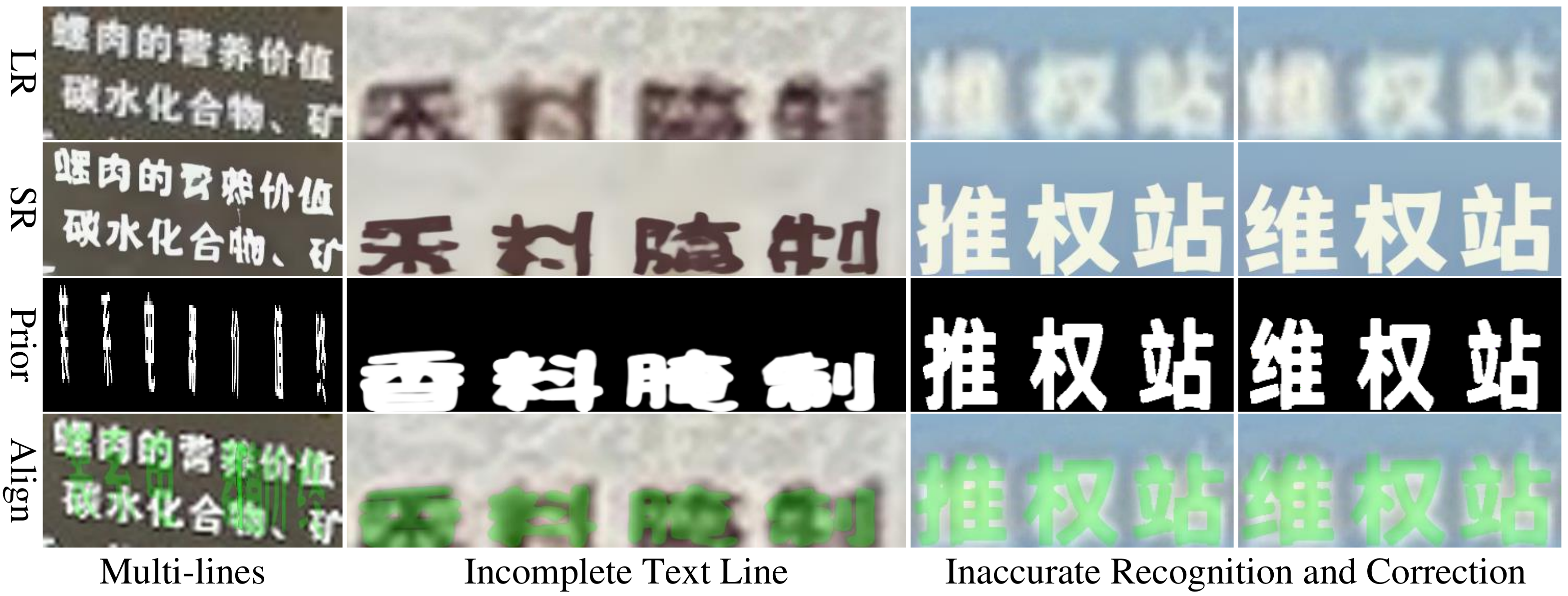}
    \caption{Failure cases of our method in real-world scenarios, including multiple text lines, incomplete text lines, and inaccurate character recognition.}
    \label{fig:failure}
\end{figure}

\subsection{Limitations and Future Work}
Our work aims to incorporate the generative structure prior into each LR character using synthetic training data. While our method excels in document-related scenes (\eg, newspaper, invoice, and scanned paper), its performance may be limited in diverse real-world scenarios due to the domain gap between real-world and synthetic text images (see the illumination inconsistency in the SR result of the last row in Figure~\ref{fig:vis}). Addressing these limitations requires real-world text images of better quality to bridge the domain gap. Another limitation is the challenge posed by text detection and recognition in real-world LR scenes. Figure~\ref{fig:failure} shows some failure cases, which are also challenging for other text SR works. Further development efforts are necessary to improve text detection and recognition capabilities in real-world LR settings.

{Our current framework demonstrates promising results in restoring degraded Chinese characters, while also highlighting several key areas for future exploration. First, although the codebook is tailored to Chinese, many characters in Japanese and Korean share structural subcomponents (\eg, radicals) with Chinese. A potential generalization strategy is to develop a hybrid cross-lingual codebook that incorporates shared radicals or stroke groups across CJK languages. This could enable broader generalization and structure prior transfer without retraining, especially for non-Chinese or open-set characters. Constructing such a codebook would involve decomposing characters into common units and clustering them into a unified representation space. However, challenges may remain in spatially composing these units under severe degradation, especially for generating stroke-level aligned structural priors.
}

{
Second, while our degradation pipeline includes various types of blur, noise, and compression artifacts from BSRGAN~\cite{zhang2021designing} and Real-ESRGAN~\cite{wang2021realesrgan}, it does not explicitly simulate motion blur, a frequent issue in real-world scenarios. 
Our method shows promising robustness against motion blur due to its strong structure prior (see Figure~\ref{fig:motion}), which helps resolve stroke-level ambiguities even under motion degradation. However, as highlighted by the red boxes, the model’s ability to recover accurate character structures is still limited under severe motion blur, where both local strokes are heavily distorted. Future work will consider motion blur-aware degradation modeling during training and explore modules specifically designed to handle motion-induced ambiguities in stroke layout and alignment.
}

\section{Conclusion}
\label{sec:sec5}
In this work, we made the first attempt to tackle the challenging task of restoring Chinese text images with complex structures and irregular layouts.
To address this issue,  we propose embedding a generative structural prior for each LR character.
By leveraging a codebook to store unique character-specific codes and adapting StyleGAN to control font styles, we effectively addressed challenges posed by complicated structures, diverse font styles, and varying layouts.
We have shown that such a structure prior is beneficial for restoring LR text images with faithful structures, even in cases of severe degradation and irregular arrangements.
Furthermore, our work can be potentially extended to other text-related applications, \eg, text image completion for ancient documents, font style transfer, and few-shot font generation.

\bibliographystyle{IEEEtran}
\bibliography{IEEEabrv,egbib}

\end{document}